\documentclass[10pt,twocolumn,letterpaper]{article}

\usepackage{graphicx}
\usepackage{subcaption}
\usepackage{float}
\usepackage[justification=raggedright]{caption}	
\usepackage{lscape}                                         

\usepackage[symbol]{footmisc}
\renewcommand{\thefootnote}{\fnsymbol{footnote}}

\usepackage[lined,ruled,linesnumbered]{algorithm2e}
\usepackage{animate} 

\usepackage{booktabs}                   
\usepackage{multirow}

\usepackage{makecell}

\usepackage{paralist}
\usepackage{enumitem}

\usepackage[export]{adjustbox}

\usepackage{bm}                          
\usepackage{epsfig}                      
\usepackage{graphicx}                  
\usepackage{times}
\usepackage{mathptmx}
\usepackage{mathtools}
\usepackage{amssymb,amsmath}   
\usepackage{breqn}

\usepackage{units}
\usepackage{color}

\usepackage{comment}

\usepackage{url}  
\usepackage[pagebackref=true,breaklinks=true,letterpaper=true,colorlinks,bookmarks=false]{hyperref}
\usepackage{xspace}
\usepackage[table]{xcolor}
\usepackage{setspace}
\usepackage{grfext}
\PrependGraphicsExtensions*{.jpg,.png,.PNG}

\usepackage[T1]{fontenc}

\DeclareMathAlphabet{\altmathcal}{OMS}{cmsy}{m}{n}

\newlength\paramargin
\newlength\figmargin

\newlength\secmargin
\newlength\figcapmargin
\newlength\tabcapmargin

\newcommand{\mpage}[2]
{
\begin{minipage}{#1\linewidth}\centering
#2
\end{minipage}
}

\newcommand{\topic}[1]
{
\vspace{1.5mm}\noindent\textbf{#1}
}

\long\def\ignorethis#1{}
\newcommand {\jiabin}[1]{{\color{cyan}\textbf{Jia-Bin: }#1}\normalfont}
\newcommand {\johannes}[1]{{\color{red}\textbf{Johannes: }#1}\normalfont}

\newcommand {\changil}[1]{{\color{blue}\textbf{Changil: }#1}\normalfont}

\newcommand{\tb}[1]{\textbf{#1}}

\newbox\jsavebox%

\makeatletter
\newcommand{\providelength}[1]{%
  \@ifundefined{\expandafter\@gobble\string#1}
   {
    \typeout{\string\providelength: making new length \string#1}%
    \newlength{#1}%
   }
   {
    \sdaau@checkforlength{#1}%
   }%
}

\def\xi{\mathbf{x}_i}

\graphicspath{{figure}, {example}}

\usepackage{cvpr}
\usepackage{animate}
\usepackage{subfiles} 
\cvprfinalcopy 

\def\cvprPaperID{3892} 

\ifcvprfinal\fi

\begin{document}

\title{Space-time Neural Irradiance Fields for Free-Viewpoint Video}

\author{
Wenqi Xian$^*$\\
Cornell Tech
\and
Jia-Bin Huang\\
Virginia Tech
\and
Johannes Kopf\\
Facebook
\and
Changil Kim\\
Facebook
}

\twocolumn[{
\renewcommand\twocolumn[1][]{#1}
\maketitle
\centering \vspace{-8mm}\url{https://video-nerf.github.io} 
\newcommand{\bracketbox}[2]{%
\begin{minipage}[t]{#1}\centering%
$\underbracket[1pt][0.5mm]{\hspace{\dimexpr(#1-0.5cm)}}_{\substack{\vspace{-3.0mm}\\\colorbox{white}{~#2~}}}$%
\end{minipage}}%

\begin{center}
\newlength{\imgwidth}
\setlength{\imgwidth}{0.16\linewidth}
\includegraphics[width=\linewidth]{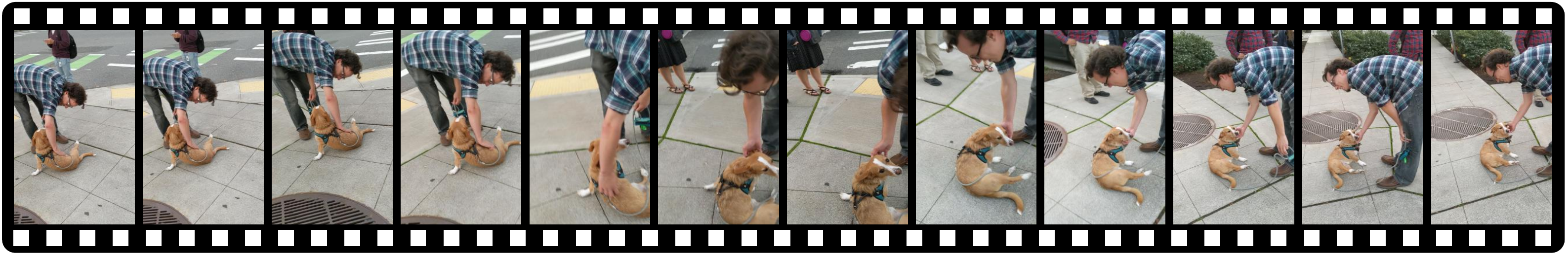}
\\[0.3mm]
\fbox{\includegraphics[trim=0 120 0 0, clip, width=\imgwidth]{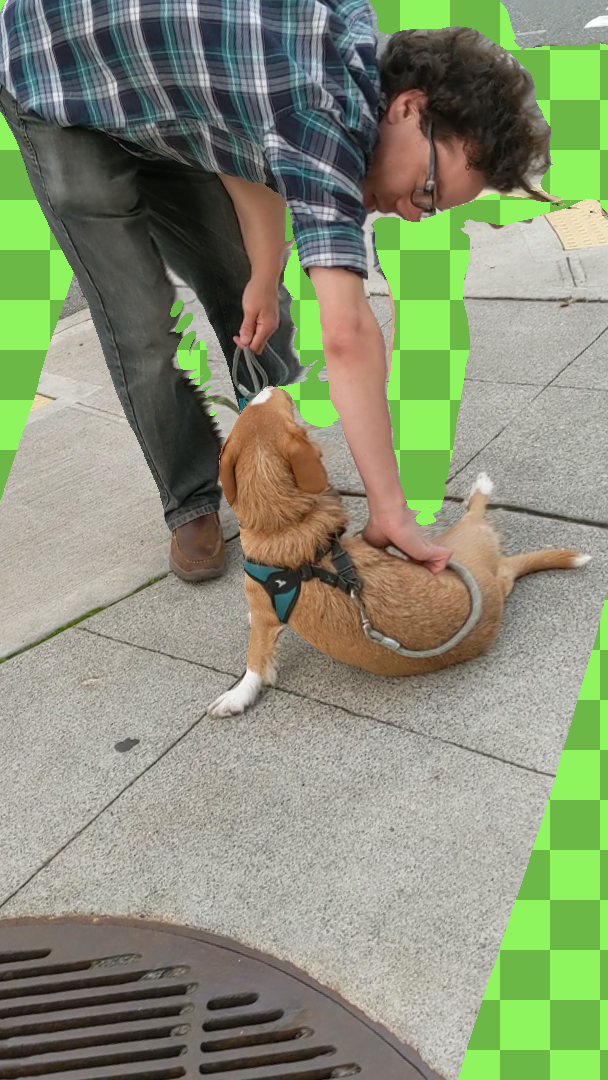}} \hfill
\fbox{\includegraphics[trim=0 120 0 0, clip, width=\imgwidth]{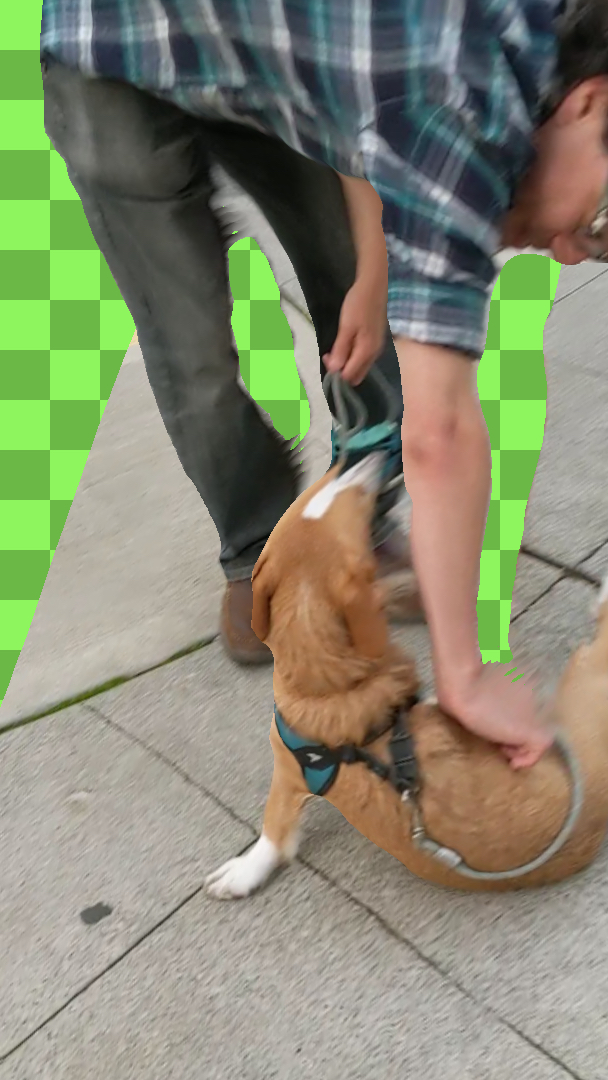}} \hfill
\fbox{\includegraphics[trim=0 120 0 0, clip, width=\imgwidth]{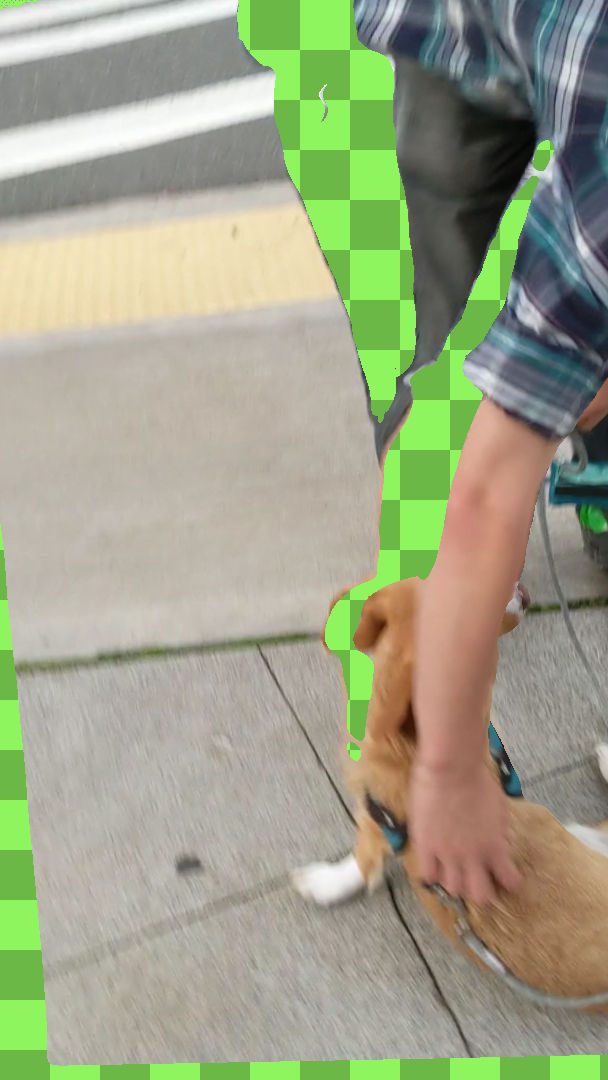}} \hfill
\fbox{\includegraphics[trim=0 120 0 0, clip, width=\imgwidth]{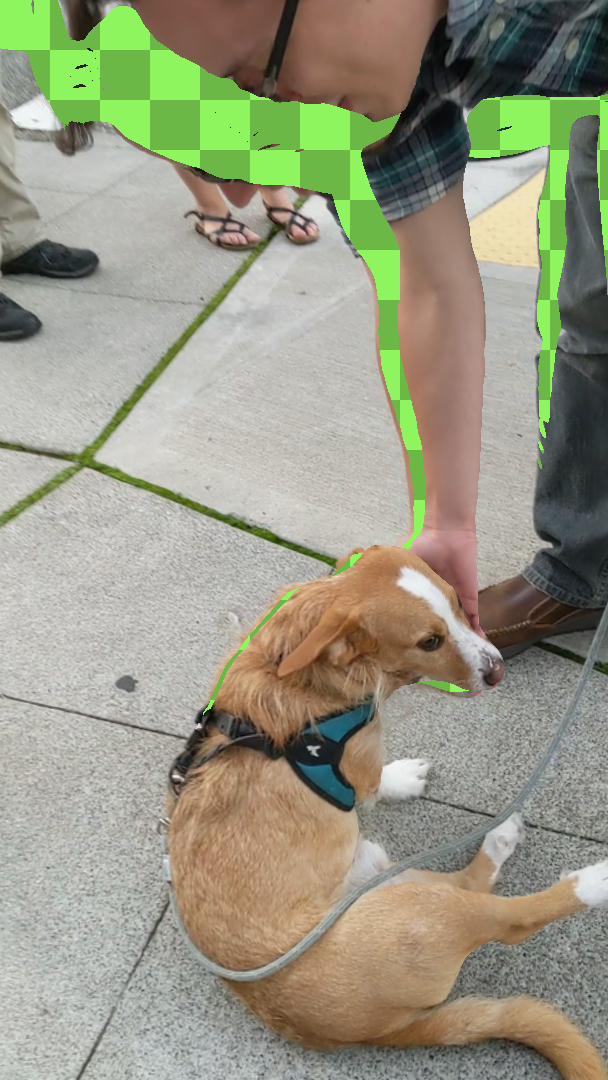}} \hfill
\fbox{\includegraphics[trim=0 120 0 0, clip, width=\imgwidth]{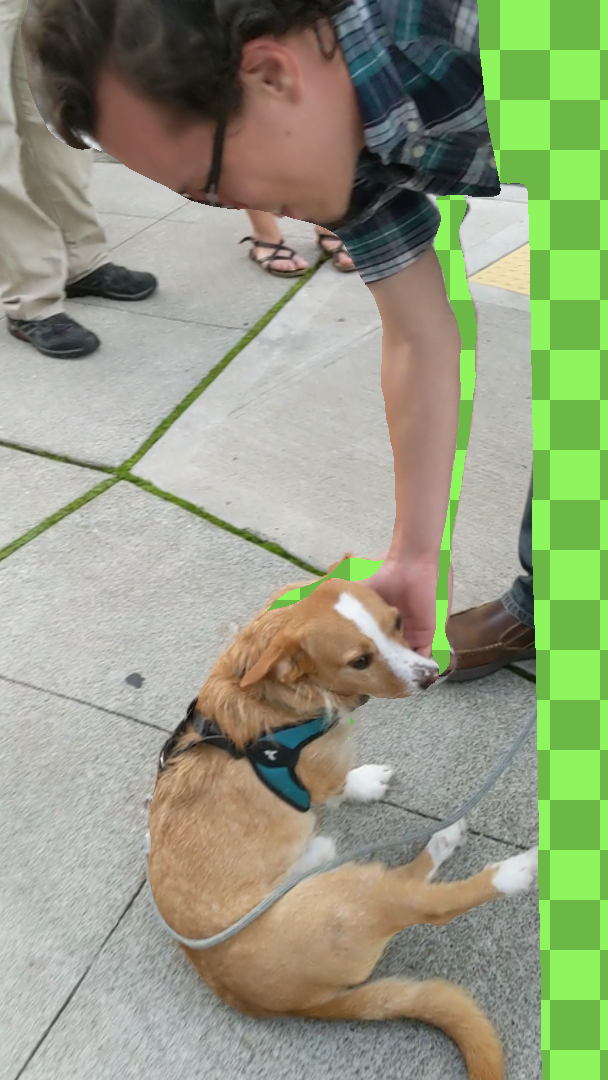}} \hfill
\fbox{\includegraphics[trim=0 120 0 0, clip, width=\imgwidth]{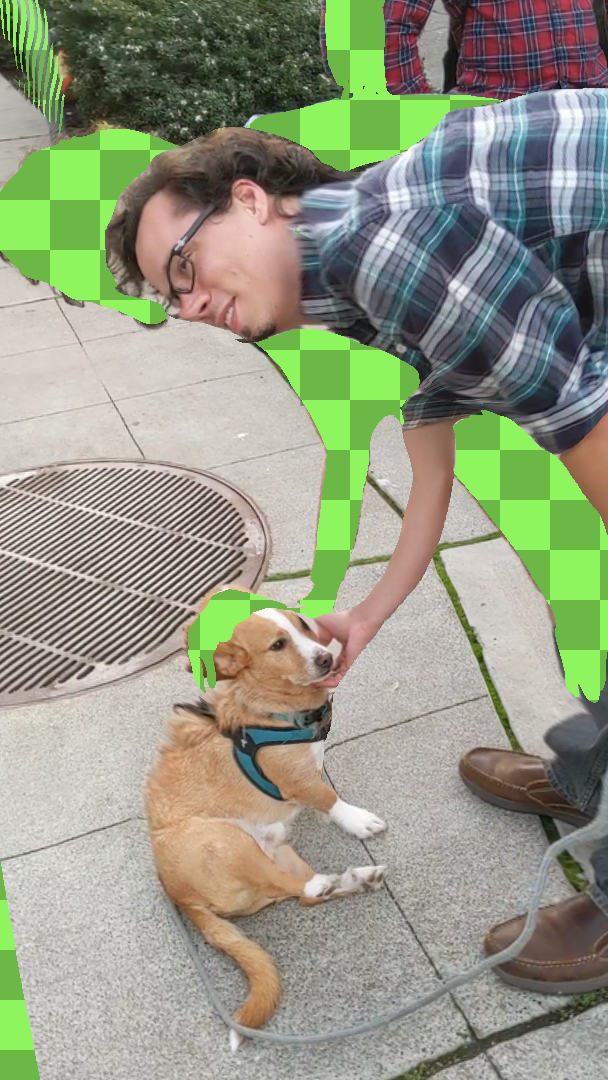}}
\\[0.7mm]
\fbox{\includegraphics[trim=0 120 0 0, clip, width=\imgwidth]{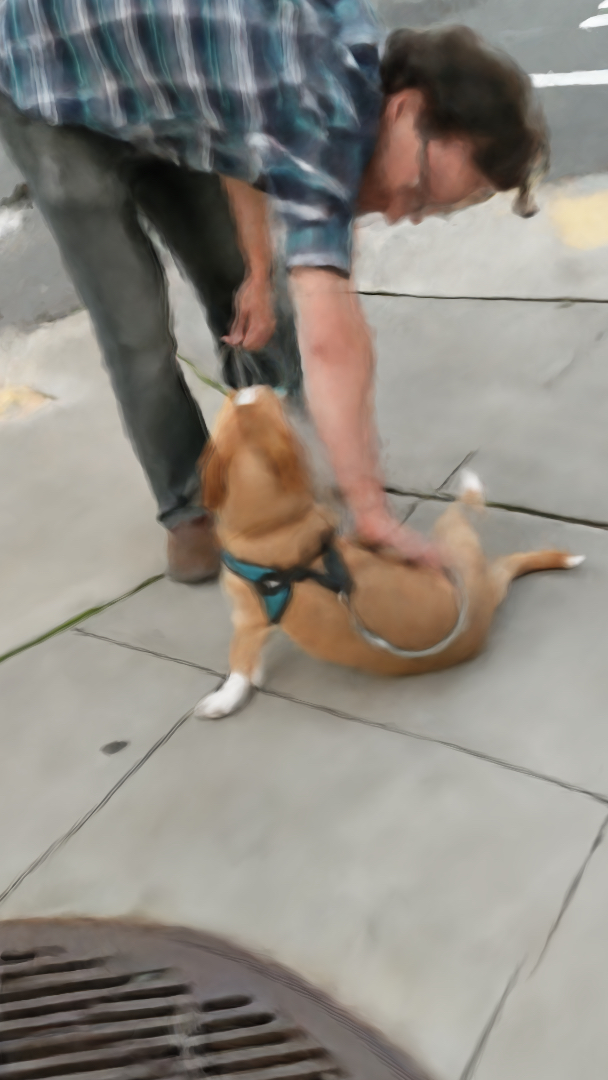}} \hfill
\fbox{\includegraphics[trim=0 120 0 0, clip, width=\imgwidth]{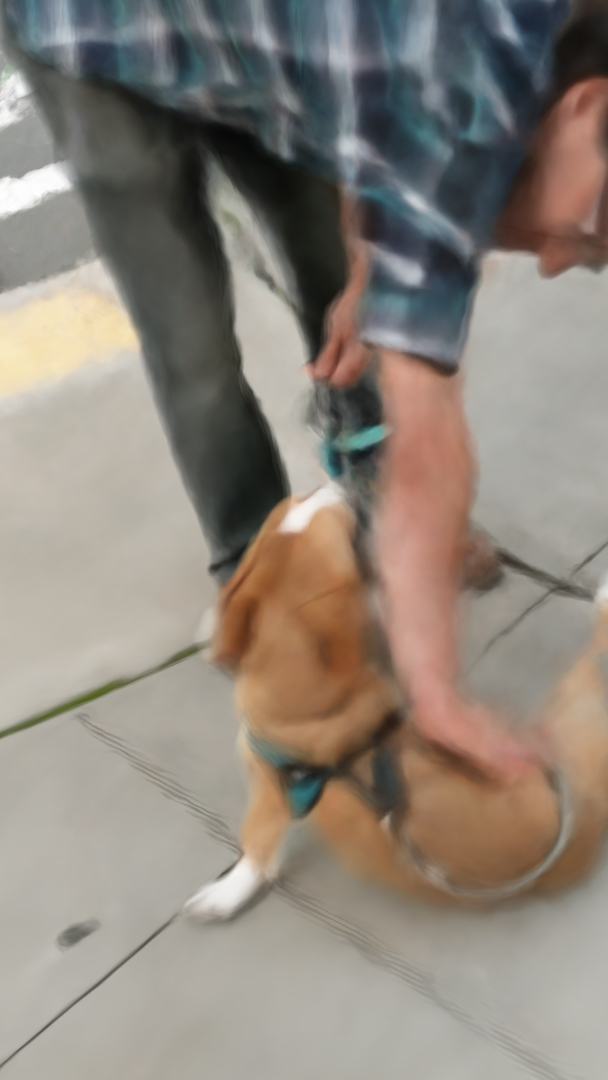}} \hfill
\fbox{\includegraphics[trim=0 120 0 0, clip, width=\imgwidth]{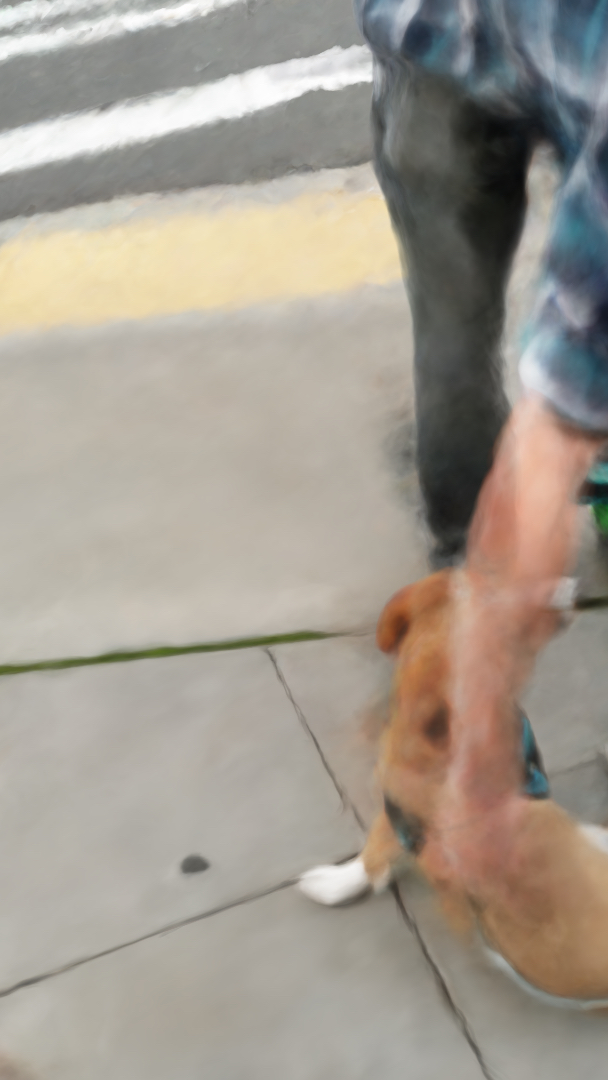}} \hfill
\fbox{\includegraphics[trim=0 120 0 0, clip, width=\imgwidth]{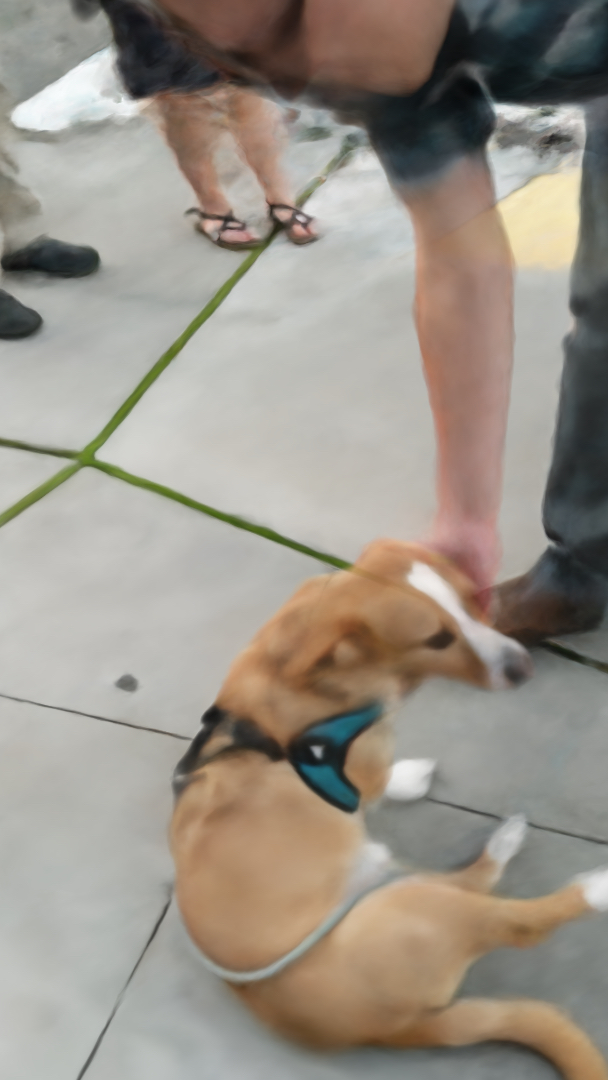}} \hfill
\fbox{\includegraphics[trim=0 120 0 0, clip, width=\imgwidth]{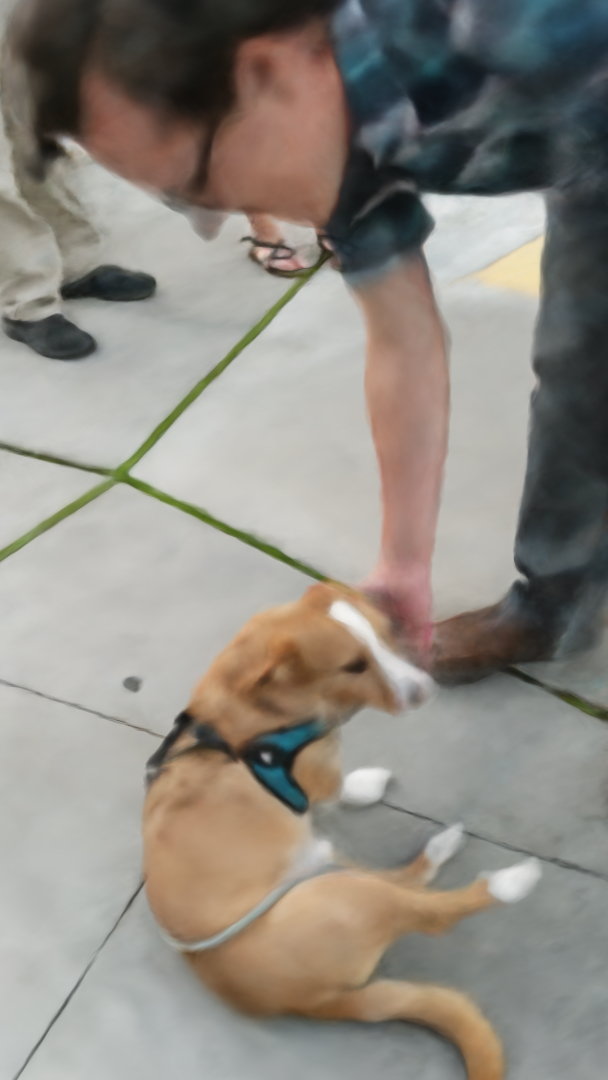}} \hfill
\fbox{\includegraphics[trim=0 120 0 0, clip, width=\imgwidth]{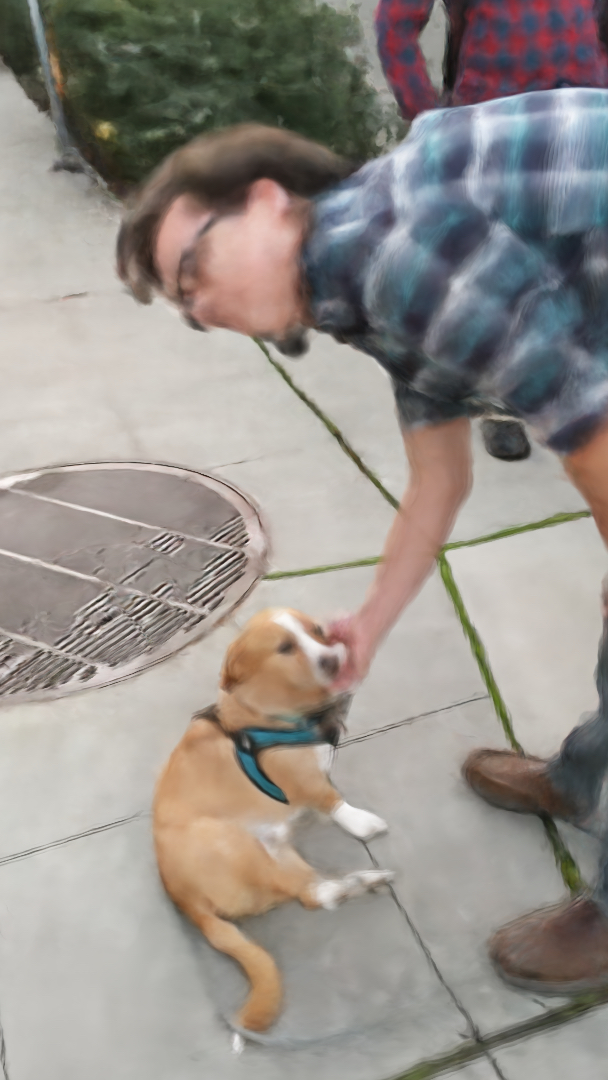}}
%
\captionof{figure}{
Our method takes a \emph{single} casually captured video as input and learns a space-time neural irradiance field. 
(\emph{Top}) Sample frames from the input video.
(\emph{Middle}) Novel view images rendered from textured meshes constructed from depth maps.
(\emph{Bottom}) Our results rendered from the proposed space-time neural irradiance field.
}
\end{center}


%

}]

\thispagestyle{empty}
\begin{abstract}
\noindent
{\let\thefootnote\relax\footnote{{$^*$ This work was done while Wenqi was an intern at Facebook.}}}
We present a method that learns a spatiotemporal neural irradiance field for dynamic scenes from a single video. Our learned representation enables free-viewpoint rendering of the input video. Our method builds upon recent advances in implicit representations. Learning a spatiotemporal irradiance field from a single video poses significant challenges because the video contains only one observation of the scene at any point in time. The 3D geometry of a scene can be legitimately represented in numerous ways since varying geometry (motion) can be explained with varying appearance and vice versa. We address this ambiguity by constraining the time-varying geometry of our dynamic scene representation using the scene depth estimated from video depth estimation methods, aggregating contents from individual frames into a single global representation. We provide an extensive quantitative evaluation and demonstrate compelling free-viewpoint rendering results.

\end{abstract}

\section{Introduction}
\label{sec:intro}

\noindent
This paper addresses the problem of rendering a video from novel viewpoints.
Specifically, we learn a \emph{globally consistent, dynamic} scene representation that can later be rendered from a novel viewpoint.
We learn such a representation from a casually captured \emph{single} video from everyday devices such as smartphones, without the assistance of multi-camera rigs or other dedicated hardware (which are typically not accessible to casual users).

Free-viewpoint video rendering typically requires a complicated hardware setup consisting of multiple cameras to capture the scene of interest from different viewpoints~\cite{zitnick2004high,collet2015high,dou2016fusion4d,broxton2020immersive}.
The multi-camera setup in existing methods is required because conventional 3D reconstruction algorithms (multi-view stereopsis) assume a fully \emph{static} scene and thus can perform reconstruction using the multiple captured viewpoints of a dynamic scene at any time.
Most methods represent the geometric reconstructions as some form of \emph{per-frame} representation (e.g., depth maps~\cite{zitnick2004high} or meshes\cite{broxton2020immersive}).
Rendering from a novel viewpoint can then be achieved, e.g., by warping available views using their depth maps to the new viewpoint.

Following the success of single-image depth estimation, recent monocular video depth estimation methods allow for the acquisition of consistent per-frame depth estimates from only a single video~\cite{Luo-VideoDepth-2020,Yoon-2020-CVPR}. 
While still at an early stage, this line of work opens up new possibilities, where monocular scene depth estimates can be directly used for view synthesis. 
However, na\"ive approaches such as per-frame depth-based warping would lead to unnatural stretches and reveal holes in disoccluded regions (even with perfect depth estimates).
One can alleviate this by post-processing the incomplete rendering~\cite{Yoon-2020-CVPR}.
However, such per-frame processing methods often lead to temporal flickers.
The core problem lies in the use of a \emph{frame-wise representation} (e.g., depth maps associated with the input images), and therefore suffer from issues ranging from temporal inconsistency to high redundancy and thus excessive storage requirements and data transfer bandwidth.

In this work, we build on recent monocular video depth estimation methods and aggregate the entire spatiotemporal aspects of a dynamic scene in a single global representation.
While fusing multiple depth maps into a single, global representation has a long tradition, most work on volumetric depth integration has focused on static scenes~\cite{Curless96,Newcombe11} or geometry alone without textures~\cite{newcombe2015dynamicfusion}.
These methods typically use \emph{discrete} representations such as voxel grids, meshes, or point clouds. Consequently, these methods often suffer from premature hard decisions on geometry estimation and limited resolution due to high storage requirements.

In this paper, we, instead, turn to the recent advances in neural implicit representations, which allow for \emph{continuous} representations of a scene without resolution loss.
Recent work has shown that these representations achieve high-quality view interpolation of complex \emph{static} scenes while retaining their advantages over discrete representations~\cite{mildenhall2020nerf,zhang2020nerf++,liu2020neural}.
The current approaches to learn them, however, require either multiple posed images of a \emph{fully static} scene~\cite{mildenhall2020nerf,zhang2020nerf++,liu2020neural} 
or ground truth 3D representations~\cite{saito2019pifu,saito2020pifuhd}. 
While videos often contain appearances of a scene seen from multiple viewpoints, they only contain \emph{exactly one} viewpoint at \emph{any given time}. 
Combined with the dynamic nature of video, this renders it nontrivial to extend current approaches to learn spatiotemporal representations from a single video. 

Specifically, we learn neural irradiance fields as a function of both space and time for each video. We do not model view dependency, hence we use the term irradiance.
Using supervision from only color frames of the input video as in \cite{mildenhall2020nerf} is futile, since the variations between frames can be explained with \emph{either} a change of appearance \emph{or} geometry, or a \emph{combination} of both. 
We resolve this ambiguity using the per-frame scene depth estimated from monocular video depth estimation.
Our depth supervision constrains the scene's geometry at any moment and disambiguates it from appearance variations.
While this enables us to encode physically correct appearance and geometry in a global representation, it fails to fill the holes that could be seen at other time steps in the video. 
We address this by encouraging the color and volume density to propagate across time whenever spatial locations are not supervised otherwise.
The resulting representation allows us to render the video from novel viewpoints and time:
our implicit model can be queried at any spatiotemporal location and rendered using standard volume rendering\footnote{Note that our method can render arbitrary viewpoints at all the \emph{observed} time steps. We do not extrapolate or interpolate the time steps.}.


Our technical contributions include the following:
\begin{itemize}
    \item We aggregate frame-wise 2.5D representations 
    into a globally consistent spatiotemporal representation from a single monocular video.
    \item We address the inherent motion--appearance ambiguity using video depth supervision and constrain the disoccluded contents by propagating the color and volume density across time. 
    \item We demonstrate a compelling free-viewpoint video rendering experience on various casual videos shot from smartphones, preserving motion and texture details while conveying a vivid sense of 3D.
\end{itemize}

\section{Related Work}
\label{sec:related}

\topic{View synthesis for images.}
Creating novel views from multiple images is a long-standing problem in computer vision and computer graphics.
Existing image-based rendering techniques first extract approximated geometric proxy and create novel target views by warping and blending the corresponding contents from multiple source frames~\cite{kalantari2016learning,penner2017soft,hedman2018deep,riegler2020free}. Multi-plane images (MPIs) have been popular in the past few years as a geometric representation~\cite{zhou2018stereo,mildenhall2019local,flynn2019deepview,li2020crowdsampling,huang2020semantic}, allowing for compelling novel view synthesis quality.
Many recent works further push the requirement of the number of input images to a narrow-baseline stereo pair~\cite{zhou2018stereo,srinivasan2019pushing,choi2019extreme} or a single image~\cite{zhou2016view,tucker2020single,niklaus20193d,wiles2020synsin,kopf2020one,shih20203d}.
Recently, neural implicit representation has shown high-quality view synthesis results by implicitly modeling the volume density and color of the scene using the weights of a multi-layer perceptron~\cite{mildenhall2020nerf,zhang2020nerf++,liu2020neural}. 
Our work uses NeRF~\cite{mildenhall2020nerf} as our base scene representation for view synthesis.
Unlike NeRF that only models \emph{static} scenes, our focus is on creating new views from arbitrary viewpoint and time for \emph{dynamic} scenes.





\topic{View synthesis for videos.}
Compared to images, view synthesis for video poses significant challenges due to the need to handle \emph{time-varying} scene geometry and appearances.
Consequently, most of the existing methods typical require laborious multi-camera setup~\cite{zitnick2004high,collet2015high,orts2016holoportation,dkabala2016efficient,broxton2020immersive}, special hardware~\cite{attal2020matryodshka}, or synchronous video captures from multiple viewpoints~\cite{ballan2010unstructured,bansal20204d}.
Several methods can reduce the required number of input views by focusing on specific domains such as performance capture~\cite{carranza2003free,dou2016fusion4d,habermann2019livecap} and video re-animation~\cite{kim2018deep,liu2020neural,liu2018neural,shysheya2019textured,chan2019everybody}.
In contrast, our work aims to enable view synthesis of a \emph{complex dynamic scene} at any given viewpoints and time from a \emph{single} video. 
Very recently, Yoon et al. \cite{Yoon-2020-CVPR} also explore the same problem setup. 
In contrast to \cite{Yoon-2020-CVPR} that processes each novel view \emph{independently} (by warping blending multiple images), we can render an interpolation video with temporally smooth transitions across viewpoints. Our method does not assume a simple two-layer (foreground-background) model of the scene and can handle more generic scenes.















\topic{Neural implicit representation.}
Implicit representation has emerged as a powerful tool to overcome conventional limitations of \emph{discrete} 3D representations such as voxel grids or meshes. 
The core idea is to use a multilayer perceptron (MLPs) to implicitly model the occupancy~\cite{mescheder2019occupancy,michalkiewicz2019implicit}, signed distance functions~\cite{park2019deepsdf,atzmon2020sal}, object appearance~\cite{oechsle2019texture}, volumetric density~\cite{eslami2018neural, lombardi2019neural, sitzmann2019deepvoxels,thies2019deferred, sitzmann2019scene,mildenhall2020nerf} in the 3D space. 
Differentiable rendering techniques enable training these models without accessing ground truth data for direct 3D supervision~\cite{mantiuk2020state,Niemeyer2020DifferentiableVR,mildenhall2020nerf,liu2020neuralvoxel}.
However, extending the above methods to handle scene dynamics is not trivial due to the motion--appearance ambiguity.
Occupancy flow~\cite{niemeyer2019occupancy} achieves 4D reconstruction (3D shape + 1D time) by learning continuous motion fields. 
Our work builds upon the recent advances in neural implicit representation but focuses on representing a \emph{dynamic video}.
Compared to 4D reconstruction in \cite{niemeyer2019occupancy}, our method differs in the following two aspects. First, our method does not require direct 3D ground truth training data. Second, in addition to model the time-varying 3D geometry, we also model the appearance of complex scenes.








\topic{Video depth estimation.}
Estimating dense depth from a dynamic video is a challenging task. 
Existing multi-view stereo (MVS) methods (either geometric-based~\cite{schonberger2016structure,furukawa2015multi} or learning-based~\cite{huang2018deepmvs,yao2018mvsnet,gu2020cascade,long2020occlusion}) assume static scene and thus not suitable for dynamic videos as it often produces erroneous depth for moving objects or untextured regions.
Several monocular video depth estimation methods predict depth using the cost volume computed from nearby frames \cite{teed2020deepv2d,liu2019neural}.
Similar to MVS algorithms, these video-to-depth methods have difficulties in handling dynamic scenes well.
Very recently, hybrid methods that combine MVS and single-image depth estimation models have been proposed \cite{Luo-VideoDepth-2020,Yoon-2020-CVPR,kopf2020robust}.
Our work leverages the estimated depth from ~\cite{Luo-VideoDepth-2020} to help resolve the ambiguity when learning the spatiotemporal neural radiance fields using a single video. 
Our approach renders photorealistic views with correctly filled dis-occluded contents compared to view synthesis with per-frame depth-based warping.


\topic{Video completion.} 
State-of-the-art video completion methods achieve temporally consistent completion by propagating known contents to missing regions along flow trajectories \cite{ilan2015survey,huang2016temporally,xu2019deep,gao2020flow}.
One may first render a free-viewpoint video according to each frame's estimated depth, followed by filling the missing pixels (disoccluded regions due to view changes) using video completion algorithms. 
Our method also produces \emph{completed} novel views (i.e., with no missing pixels from disocclusion).
However, unlike video completion algorithms that inpaint the dis-occluded pixels in the \emph{screen space}, our approach fills in the dis-occluded content implicitly in the \emph{3D space}.
Our experiments validate that our approach produces significantly fewer artifacts than the baseline method using video completion.

\topic{Depth map fusion.} 
A line of research work focuses on fusing a sequence of RGB-Depth images in a video into a \emph{global} with voxel-, point-based, signed distance field-based representation~\cite{zollhofer2018state}.
Examples include 3D reconstruction for static scenes~\cite{niessner2013real,izadi2011kinectfusion} or dynamic objects from a single~\cite{newcombe2015dynamicfusion,innmann2016volumedeform} or RGB-D cameras~\cite{dou2016fusion4d,bozic2020deepdeform,innmann2016volumedeform,newcombe2015dynamicfusion,ye2014real,zollhofer2014real}.
Our work differs from prior 3D reconstruction methods in two aspects.
First, we do not assume a fixed, canonical 3D model as in existing dynamic 3D reconstruction methods and, therefore, can naturally handle an entire dynamic scene (as opposed to only individual objects).
Second, our approach with neural implicit representations jointly models time-varying geometry \emph{and} appearance.

\topic{Concurrent works.} Several concurrent works on extending NeRF~\cite{mildenhall2020nerf} to dynamic scenes from monocular video have been proposed~\cite{park2020deformable, pumarola2020d, li2020neural, tretschk2020non, du2020neural,gao2021dynamic}. These methods either learn a static canonical radiance field with deformation~\cite{park2020deformable, pumarola2020d, tretschk2020non} or a dynamic radiance field directly conditioned on time~\cite{li2020neural, du2020neural,gao2021dynamic}. Our work belongs to the latter; while the other two works~\cite{li2020neural, du2020neural} regularize the training primarily with flow information, ours does so with dynamic scene depth. We refer the readers to these papers for a complete picture.

\newcommand{\F}{F}
\newcommand{\X}{\mathbf{x}}
\newcommand{\Dir}{\mathbf{d}}
\newcommand{\T}{t}
\newcommand{\Tdom}{\altmathcal{T}}
\newcommand{\C}{\mathbf{c}}
\newcommand{\A}{\sigma}
\newcommand{\Loss}{\bm{\altmathcal{L}}}
\newcommand{\Lcol}{\Loss_{\mathrm{color}}}
\newcommand{\Ldep}{\Loss_{\mathrm{depth}}}
\newcommand{\Lstatic}{\Loss_{\mathrm{static}}}
\newcommand{\Lfree}{\Loss_{\mathrm{empty}}}
\newcommand{\Lflow}{\Loss_{\mathrm{flow}}}
\newcommand{\R}{\mathbf{r}}
\newcommand{\Rdom}{\altmathcal{R}}
\newcommand{\volrend}{C}
\newcommand{\deprend}{D}
\newcommand{\slack}{\epsilon}
\newcommand{\pool}{\altmathcal{X}}
\newcommand{\NumRays}{N_{\mathrm{r}}}
\newcommand{\NumStaticSamples}{N_{\mathrm{s}}}
\newcommand{\NumFrames}{N_{\mathrm{f}}}
\newcommand{\calib}{\altmathcal{P}}
\newcommand{\I}{I}
\newcommand{\U}{\mathbf{u}}
\newcommand{\Z}{z}
\newcommand{\Zn}{\Z_{\mathrm{n}}}
\newcommand{\Zf}{\Z_{\mathrm{f}}}
\newcommand{\Rad}{L}
\newcommand{\W}{W}
\newcommand{\Wfwd}{\W_{\mathrm{f}}}
\newcommand{\Wbwd}{\W_{\mathrm{b}}}
\newcommand{\Rp}{s}
\newcommand{\Rpn}{\Rp_{\mathrm{n}}}
\newcommand{\Rpf}{\Rp_{\mathrm{f}}}
\newcommand{\Rpp}{p}
\newcommand{\Ori}{\mathbf{o}}

\section{Background}

\noindent
Our representation builds on the neural radiance field, or NeRF~\cite{mildenhall2020nerf}, which we recap in this section.
NeRF represents the radiance $\C = (r, g, b)$ and differential volume density $\A$ at a 3D location $\X = (x, y, z)$ of a scene observed from a viewing direction $\Dir = (\theta, \phi)$ as a continuous multi-variate function using a multi-layer perceptron (MLP): $\F_{\mathrm{NeRF}}: (\X, \Dir) \rightarrow (\C, \A)$.

The color of a pixel can be rendered by integrating the radiance modulated by the volume density along the camera ray $\R(\Rp) = \Ori + \Rp \Dir$, shot from the camera center through the center of the pixel:
\begin{equation}
    \volrend(\R) = \int_\Rpn^\Rpf T(\Rp) \A\Big(\R(\Rp)\Big) \C\Big(\R(\Rp),\, \Dir\Big)\mathrm{d}\Rp,
\label{eq:nerf_rendering}
\end{equation}
where
\begin{equation}
    T(\Rp) = \exp \!\left( - \!\int_\Rpn^\Rp \A(\R(\Rpp))\mathrm{d}\Rpp \right)
\end{equation}
is the accumulated transmittance along the ray $\R$ up to $\Rp$.

One can train the MLP using multiple posed images, capturing a \emph{static} scene from different viewpoints.
Specifically, we minimize the photometric loss that compares the rendering through a ray $\R$ with the corresponding ground truth color from an input image:
\begin{equation}
    \Loss_{\mathrm{NeRF}} = \sum_{\R \in \Rdom} \left\| \hat{\volrend}(\R) - \volrend(\R) \right\|_2^2 ,
\label{eq:nerf_loss}
\end{equation}
where $\Rdom$ denotes a set of rays, and $\volrend(\R)$ and $\hat{\volrend}(\R)$ the ground truth and the estimated color, respectively.

In the implementation, the continuous volume rendering of~\eqref{eq:nerf_rendering} is approximated by numerical quadrature, i.e., computing the color using a finite number of sampled 3D points along a ray and calculate the summation of the radiances, weighted by the discrete transmittance.
As this weighted summation process is differentiable, the gradient can propagate backward for optimizing the MLP.
We perform the sampling in two steps. 
First, a ray is sampled uniformly in $\Rp$, and then, it is sampled with respect to the approximate transmittance so that more samples are used around surfaces in the scene. 
The two groups of samples are evaluated in separate \emph{coarse} and \emph{fine} networks, and both are used to measure the loss~\eqref{eq:nerf_loss}.

%

\section{Space-time Neural Irradiance Fields}
\label{sec:overview}

\noindent
We represent a 4D space-time irradiance field as a function that maps a spatiotemporal location $(\X, \T)$ to the emitted color and volume density, $\F: (\X, \T) \rightarrow (\C, \A)$. 
Our input video is represented as a stream of RGB-D images, $\I_\T: \U \rightarrow (\C, d)$ at discrete time steps $\T \in \Tdom = \{1, 2, ..., \NumFrames\}$, where $\U = (u, v)$ is 2D pixel coordinates, and their associated camera calibrations $\calib_\T$.

A ray $\R$ at time $\T$ can be determined by a pixel location $\U$ and the camera calibration $\calib_\T$: it marches from the camera center through the center of pixel denoted by $\U$. 
Additionally, we parameterize a ray such that the parameter $\Rp$ denotes the scene depth. 
This is achieved by setting the directional vector $\Dir$ such that its projection onto the principal ray has a unit norm in the camera space.

\topic{Color reconstruction loss.}
To learn the implicit function $\F$ from the input video $\I$, first and foremost, we constrain our representation $\F$ such that it reproduces the original video $\I$ when rendered from the original viewpoint for each frame. 
Specifically, we penalize the difference between the volume-rendered image at each time $\T$ and the corresponding input image $\I_\T$.
This amounts to the reconstruction loss of the original NeRF~\cite{mildenhall2020nerf}:
\begin{equation}
    \Lcol = \sum_{(\R \! , \, \T) \in \Rdom} \left\| \hat{\volrend} (\R, \T) - \volrend (\R, \T) \right\|^2_2 ,
\label{eq:color_loss}
\end{equation}
where $\Rdom$ is a batch of rays, each associated with a time $\T$.

Unlike NeRF, for dynamic scenes, we have to reconstruct the \emph{time-varying} scene geometry at \emph{every} time $\T$.
However, a single video contains only one observation of the scene at any point in time, rendering the estimation of scene geometry severely under-constrained.
That is, the 3D geometry of a scene can be legitimately represented in numerous (infinitely possible) ways since varying geometry can be explained with the varying appearance and vice versa.
For example, any input video can be reconstructed with a ``a flat TV'' solution (with a planar geometry with each frame texture-mapped).

\begin{figure}[t]
\begin{center}
\includegraphics[width=0.19\linewidth]{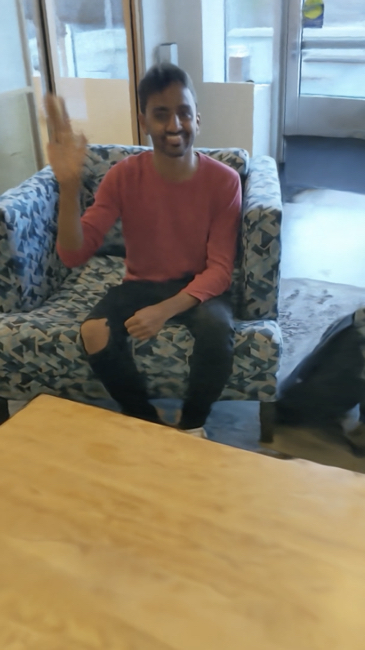}
\includegraphics[width=0.19\linewidth]{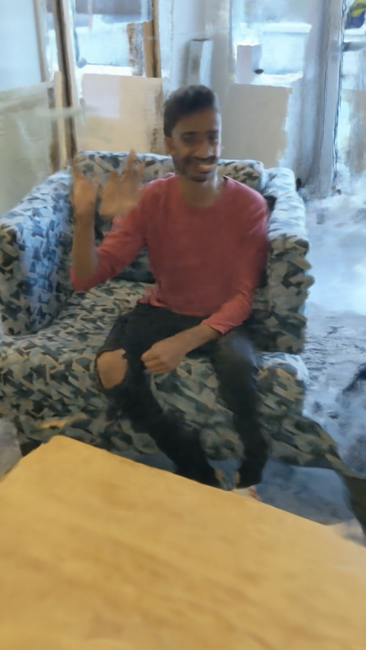}
\includegraphics[width=0.19\linewidth]{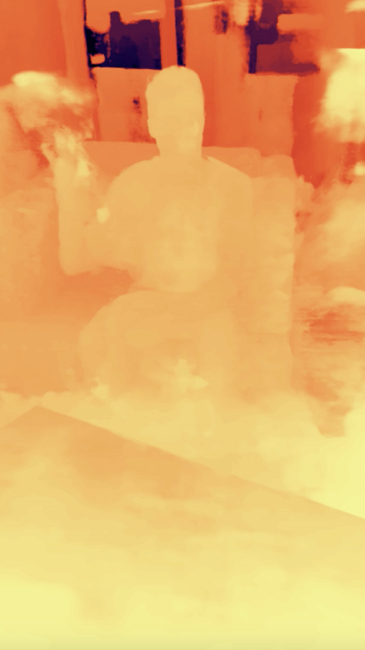} \hfill
\includegraphics[width=0.19\linewidth]{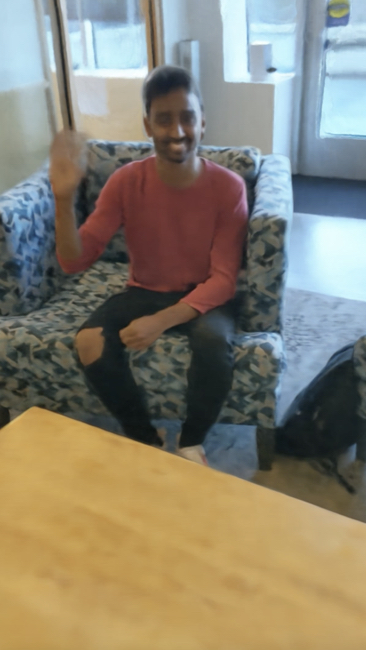}
\includegraphics[width=0.19\linewidth]{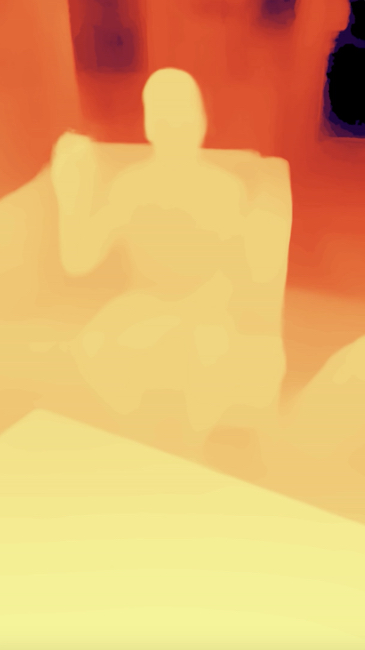}
\vspace{-2mm}
\mpage{0.58}{\small (a) w/o depth loss}
\mpage{0.38}{\small (b) w/ depth loss}
\end{center}
\caption{
\tb{Depth loss.}
(a) A model trained \emph{without} our depth loss can reconstruct the image from the original viewpoint well (\emph{first}). However, with a slight viewpoint change, the synthesized image (\emph{second}) suffers from strong visual artifacts due to incorrect geometry (\emph{third}).
(b) A trained \emph{with} the proposed depth loss render the novel viewpoint without clearly visible artifacts. 
}
\label{fig:depth_loss}
\end{figure}

Thus, the color reconstruction loss provides the ground for accurate reconstruction \emph{only} when the learned representation is rendered from the same camera trajectory of the input, lacking any machinery that drives learning correct geometry.
Incorrect geometry would lead to artifacts as soon as we start deviating from the original video's camera trajectory, as shown in Figure~\ref{fig:depth_loss}a.



\topic{Depth reconstruction loss.}
We resolve this motion--appearance ambiguity by constraining the time-varying geometry of our dynamic scene representation using the per-frame scene depth of the input video (estimated from video depth estimation methods).
We estimate the scene depth from the learned volume density of the scene, and measure its difference from the input depth $d_\T$. 
It is a non-trivial question on how to define the scene depth of a ray. 
One possibility is to measure the distance where the accumulated transmittance $T$ becomes less than a certain threshold. 
Such an approach, however, involves heuristics and hard decisions.
Instead, we accumulate depth values along the ray modulated both with the transmittance and volume density, similarly to the depth composition in layered scene representations~\cite{tucker2020single}.

Our depth reconstruction loss is of the form:
\begin{equation}
    \Ldep = \sum_{(\R \! , \, \T) \in \Rdom} \left\| \frac{1}{\hat{\deprend} (\R, \T)} - \frac{1}{\deprend (\R, \T)} \right\|^2_2,
\label{eq:depth_loss}
\end{equation}
where
\begin{equation}
    \hat{\deprend}(\R, \T) = \int_\Rpn^\Rpf T(\Rp, \T) \, \A(\R(\Rp), \T) \, \Rp \, \mathrm{d}\Rp,
\end{equation}
is the integrated sample depth values along the ray and $\deprend(\R)$.

\topic{Empty-space loss.}
Constraining the depth predicted by our model using the estimated scene depth is not sufficient to capture accurate scene geometry.
This is because the predicted depth is, in essence, a \emph{weighted sum} of depth values along the ray. 
Consequently, we sometimes see haze-like visual artifacts when rendering at novel views. 
We propose to encourage \emph{empty space} between the camera and the first visible scene surface to address this issue. 
A similar idea has been used in volumetric depth integration~\cite{Curless96}. 
To this end, we penalize non-zero volume densities measures along each ray up to the point no closer than a small margin $\slack = 0.05 \cdot (\Rpf - \Rpn)$ 
to the scene depth for each ray:
\begin{equation}
    \Lfree = \sum_{(\R \! , \, \T) \in \Rdom} \int_{\Rpn}^{d_\T(\U) - \slack} \A (\R(\Rp), \T) \, \mathrm{d}\Rp ,
\label{eq:zvd_loss}
\end{equation}
where $\U$ denotes the pixel coordinates where $\R$ intersects with the image plane at $\T$, $d_\T(\U)$ denote the scene depth for the pixel $\U$ at time $\T$.

The empty-space loss combined with the depth reconstruction loss provides geometric constraints for our representation up to and around \emph{visible} scene surfaces at each frame.
The learned representations can thus produce geometrically correct novel view synthesis, as shown in Figure~\ref{fig:depth_loss}b.

\begin{figure}[t]
\begin{center}
\includegraphics[width=\linewidth]{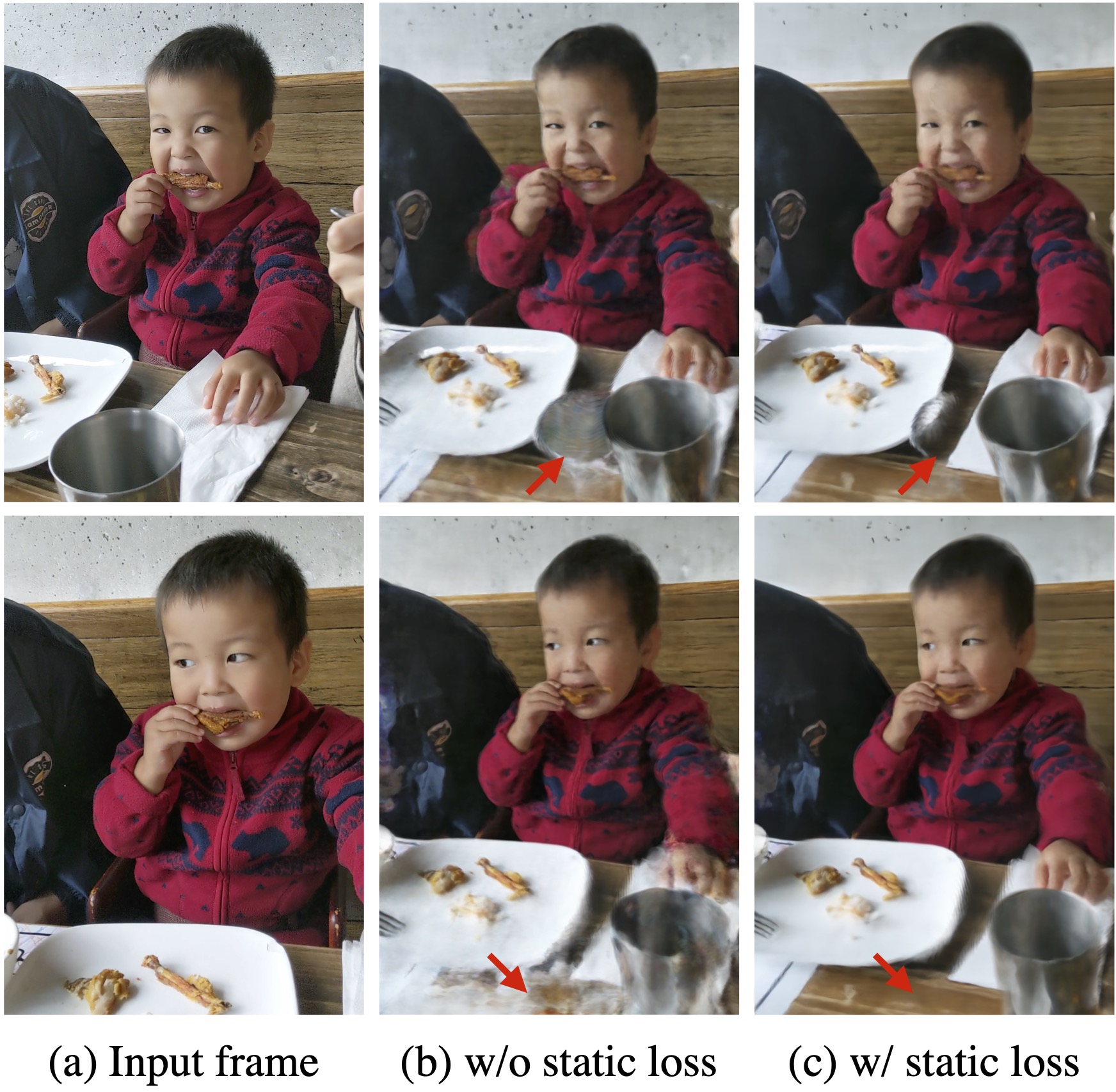}
\end{center}
\caption{%
\tb{Static scene loss.}
Our static scene loss addresses the regions that are \emph{not} constrained by the color and depth reconstruction loss, which handles only \emph{visible} surfaces in the input frames (a). 
We render two frames at $\T_1 = 9$ and $\T_2 = 90$ from the viewpoint of $\T = 40$. 
The camera has panned in the top row and zoomed out in the bottom row. 
When these frames are rendered from this novel viewpoint, previously hidden regions become \emph{disoccluded}. 
They are, without the static scene loss, completely unconstrained and prone to ghosting or haze artifacts, as shown in (b). 
Our static loss alleviate these artifacts.
}
\label{fig:static_loss}
\end{figure}

\topic{Static scene loss.}
A large portion of spaces that is \emph{hidden} from the input frame's viewpoint at any given time is still \emph{not constrained}, i.e., the MLP has not seen the 3D positions and time as input queries during training. 
As a result, when these unconstrained spaces are disoccluded due to viewpoint changes, 
they are prone to artifacts (see Figure~\ref{fig:static_loss} for an example).
However, there is a high chance that a portion of disoccluded spaces is observed from a \emph{different viewpoint} at \emph{another time}.
Our idea is to constrain the MLP by propagating these partially observed contents across time.
However, instead of explicitly correlating surfaces over time, e.g, using the scene flow, we choose to constrain the spaces surrounding the surface regions. 
This allows us to avoid misalignment of scene surfaces due to unreliable geometry estimates or other image aberrations commonly seen in captured videos such as exposure or color variations.


We make a simple assumption on \emph{unobserved} spaces: every part of the world should stay static unless observed not as such. Enforcing this assumption prevents the part of spaces that are not observed from going entirely unconstrained.
Our static scene constraint encourages the shared color and volume density at the same spatial location $\X$ between two distinct times $\T$ and $\T'$:
\begin{equation}
    \Lstatic = \sum_{(\X, \, \T) \in \pool} \left\| \F (\X, \T) - \F (\X, \T') \right\|^2_2 ,
\label{eq:static_loss}
\end{equation}
where both $(\X, \T)$ and $(\X, \T')$ are \emph{not} close to any visible surfaces, and $\pool$ denotes a set of sampling locations where the loss is measured.

\begin{figure}[t]
\begin{center}
\includegraphics[width=0.8\linewidth]{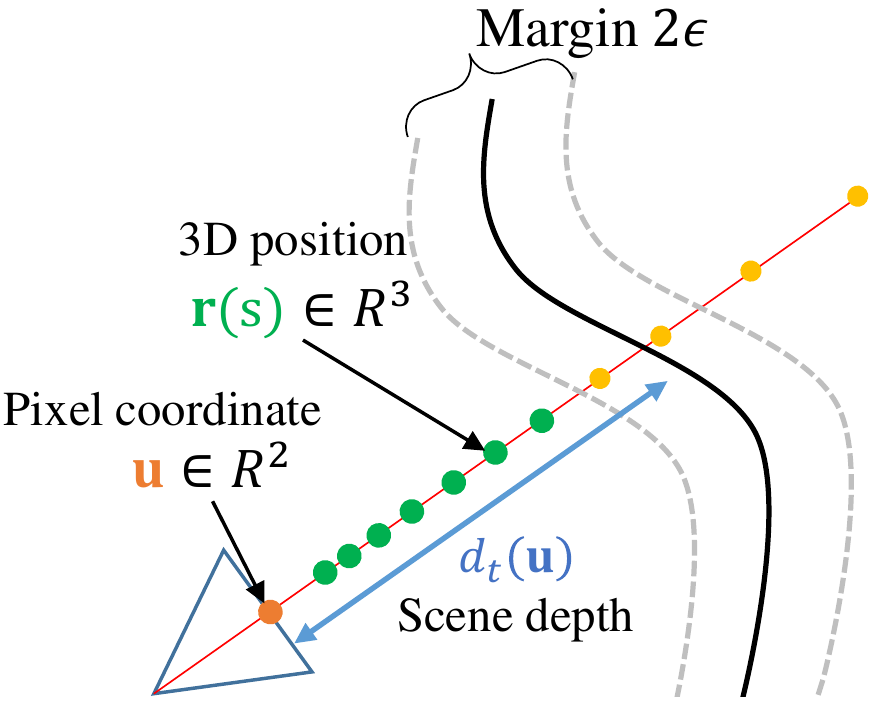}
\end{center}
\caption{
\tb{Scene sampling.}
We measure the \emph{Empty-space loss} on 3D positions until hitting the estimated scene depth (green). We use all the samples along the ray (green and yellow) to compute the \emph{Depth} and \emph{Color reconstruction loss}.
For \emph{Static loss}, we sample the scene from the union of the space spanned by all camera rays (within the range of $\Rp \in [\Zn, \Zf]$) of all input frames. We exclude any samples that are close to any surface than $\slack$. 
}
\label{fig:sampling}
\end{figure}

\topic{Scene sampling.}
While we have locations for the color, depth, and free-space supervisions explicitly dictated by quadrature used by volume rendering~\cite{mildenhall2020nerf}, we are free to choose \emph{where} we apply the static constraints.
A straightforward approach would be to use the same sampling locations that are used for other losses.
We can then randomly draw another time $\T'$ that is distinct from the current time $\T$ and enforce the MLP to produce similar appearances and volume densities at these two spatiotemporal locations.

However, this still leaves a large part of the scene unconstrained when the camera motion is large.
Uniformly sampling in the scene bounding volume would also not be ideal since sampling would be highly inefficient because of perspective projection (except for special cases like a camera circling some bounded volume).

As a simple solution to meet both the sampling efficiency and the sample coverage, 
we propose to take the \emph{union} of all sampling locations along \emph{all} rays of \emph{all} frames to form our sample pool $\pool$. 
We exclude all points that are closer to any observed surfaces than a threshold $\slack$ (see Figure~\ref{fig:sampling}).
We randomly draw a fixed number of sampling locations from this pool at each training iteration and add small random jitters to each sampling location. 
At time $\T'$ the static scene loss is measured against is also randomly chosen for each sample location $\X$, while ensuring the resulting location $(\X, \T')$ is not close to any scene surfaces.

\topic{Total loss.}
Our total loss for training the space-time irradiance fields is a linear combination of all losses presented above:
\begin{equation}
    \Loss = \Lcol + \alpha \Ldep + \beta \Lfree + \gamma \Lstatic . 
\end{equation}
We validate the effectiveness of these losses in Section~\ref{sec:results}.

\begin{figure*}[htbp!]
\begin{center}
\fbox{\adjincludegraphics[width=0.19\linewidth]{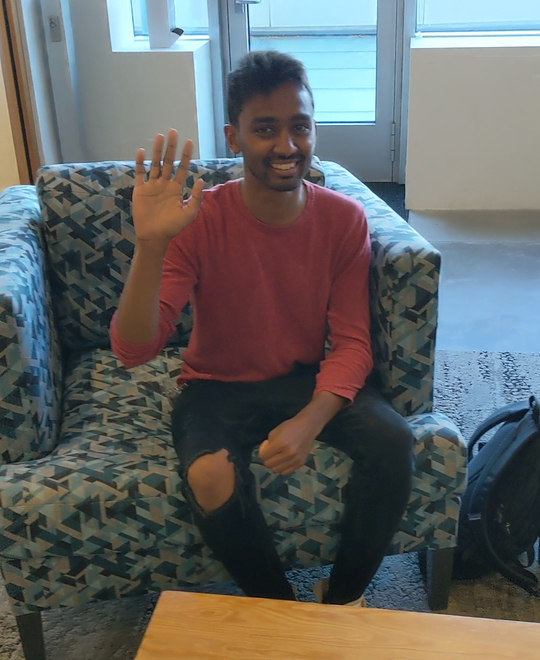}} \hfill
\fbox{\adjincludegraphics[width=0.19\linewidth]{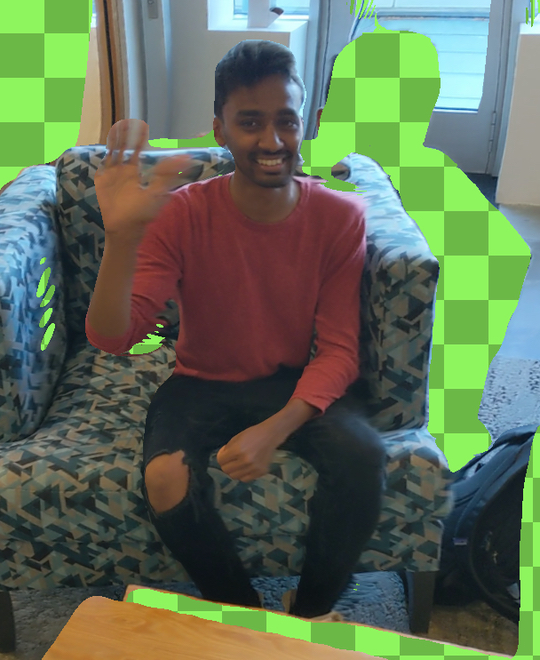}} \hfill
\fbox{\adjincludegraphics[width=0.19\linewidth]{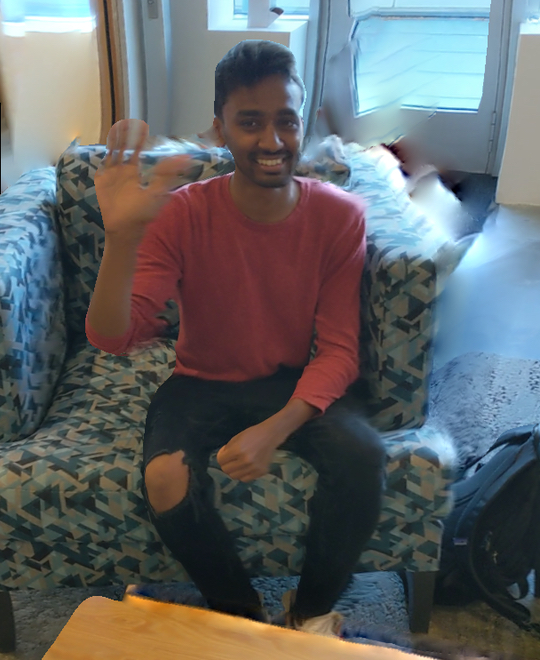}} \hfill
\fbox{\adjincludegraphics[width=0.19\linewidth]{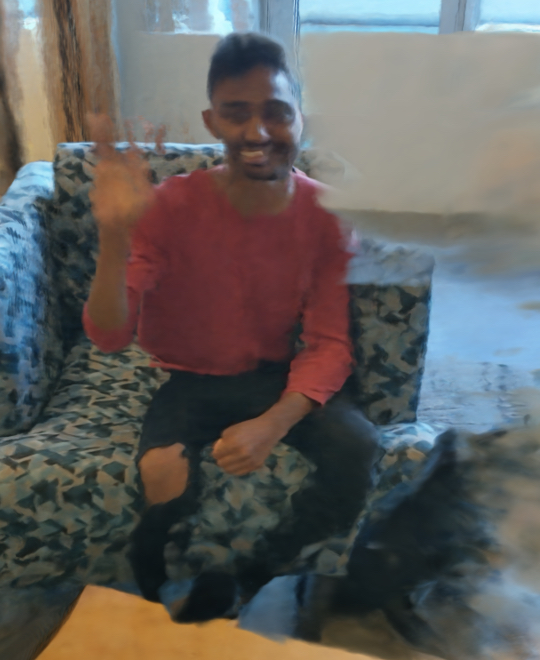}} \hfill
\fbox{\adjincludegraphics[width=0.19\linewidth]{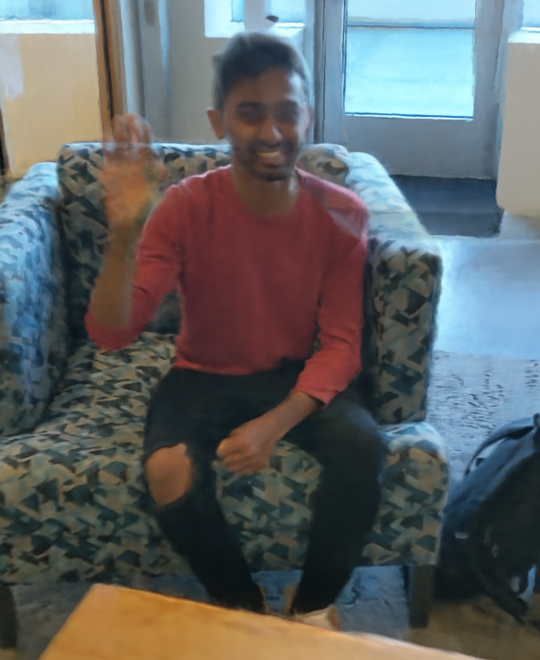}}\vspace{1mm}

\fbox{\adjincludegraphics[width=0.19\linewidth]{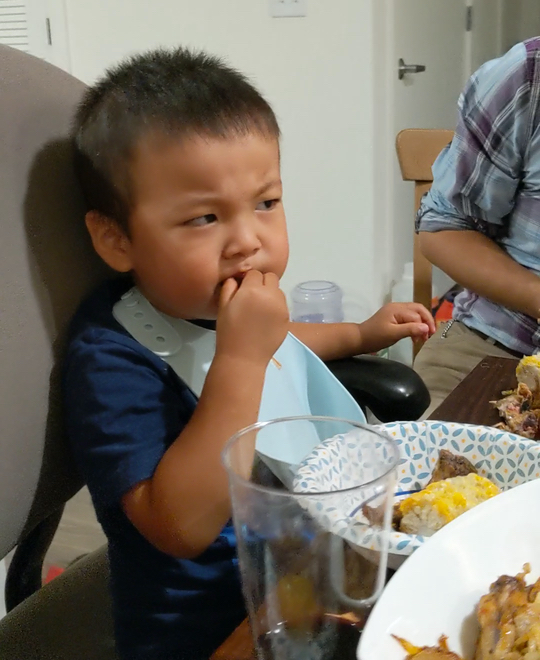}} \hfill
\fbox{\adjincludegraphics[width=0.19\linewidth]{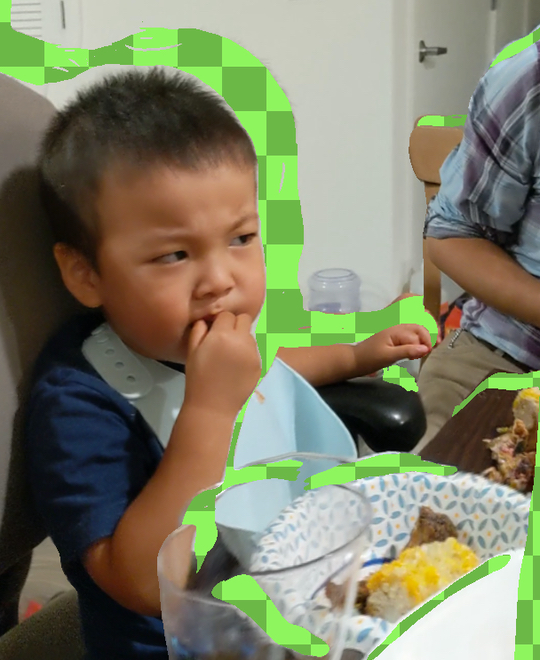}} \hfill
\fbox{\adjincludegraphics[width=0.19\linewidth]{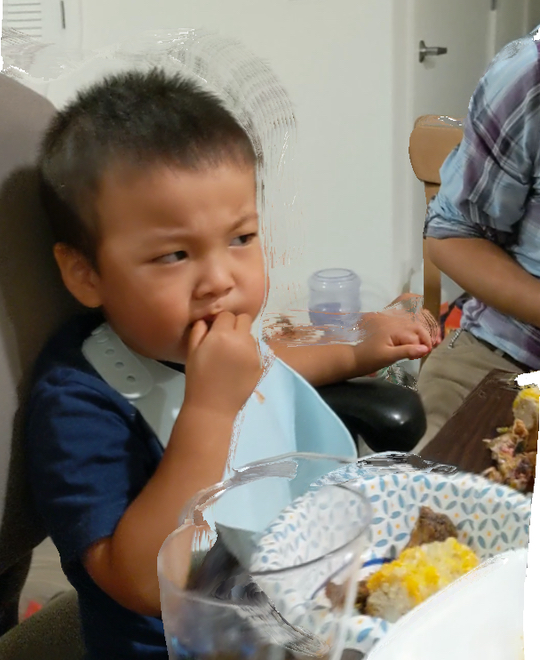}} \hfill
\fbox{\adjincludegraphics[width=0.19\linewidth]{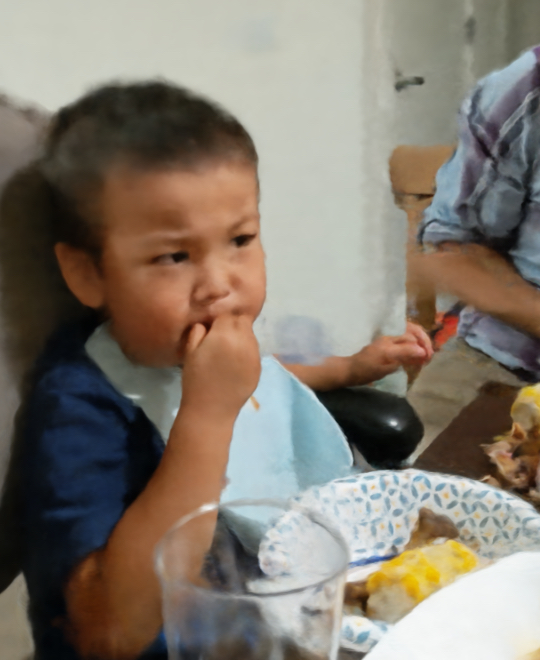}} \hfill
\fbox{\adjincludegraphics[width=0.19\linewidth]{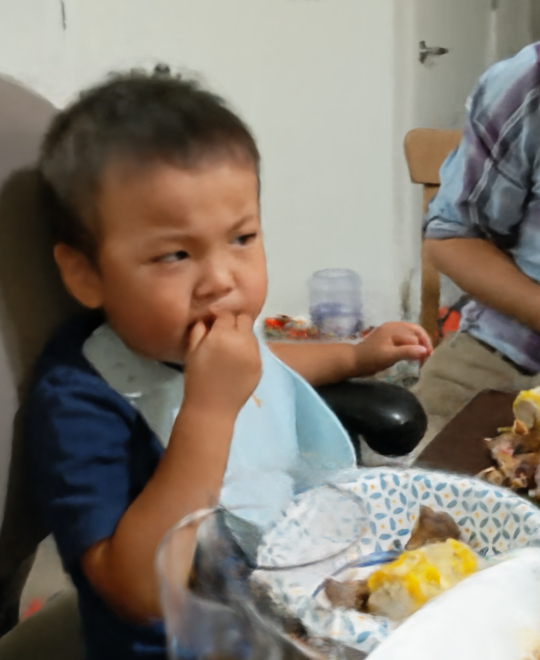}}\vspace{1mm}

\fbox{\adjincludegraphics[width=0.19\linewidth]{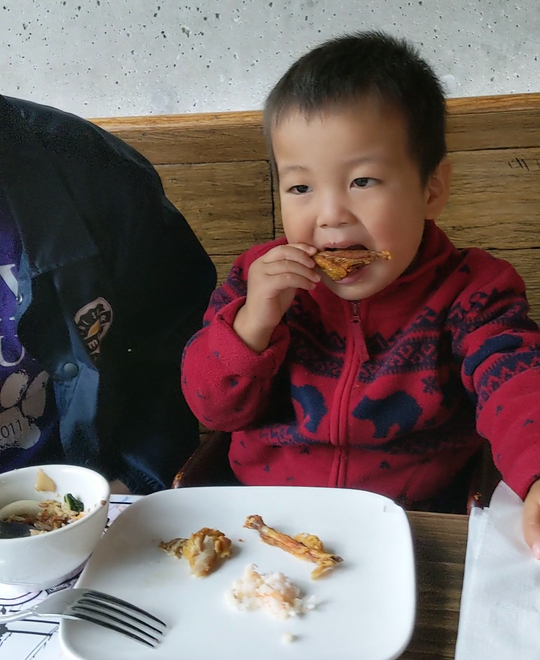}} \hfill
\fbox{\adjincludegraphics[width=0.19\linewidth]{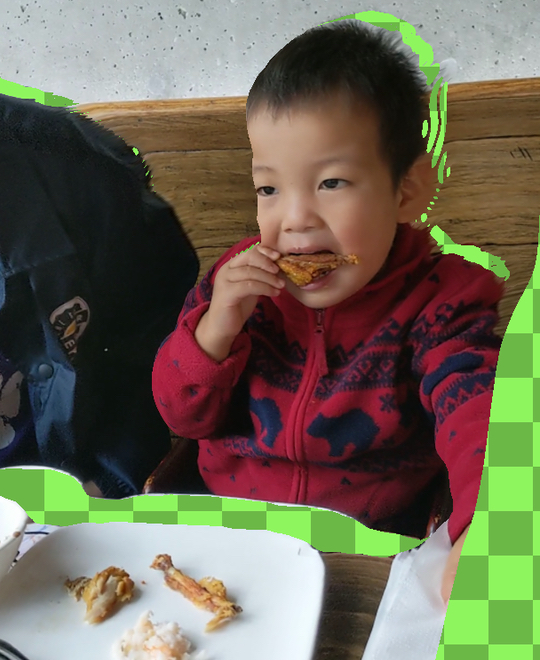}} \hfill
\fbox{\adjincludegraphics[width=0.19\linewidth]{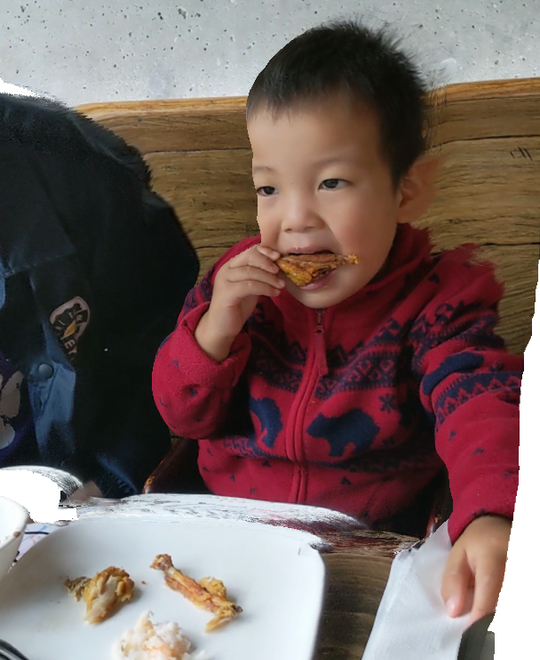}} \hfill
\fbox{\adjincludegraphics[width=0.19\linewidth]{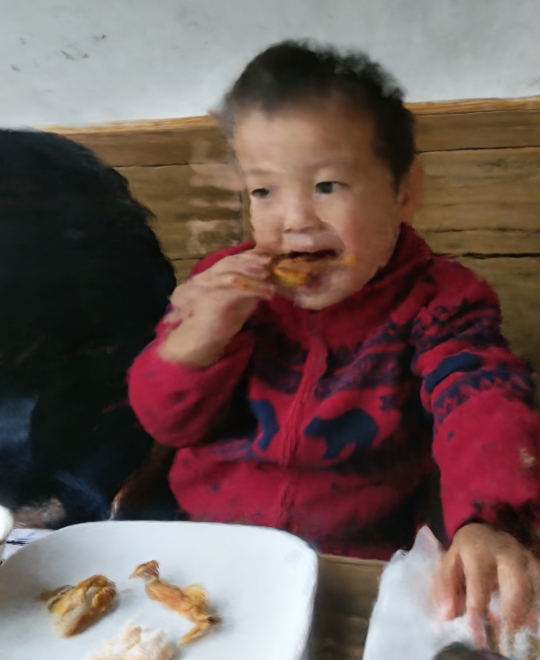}} \hfill
\fbox{\adjincludegraphics[width=0.19\linewidth]{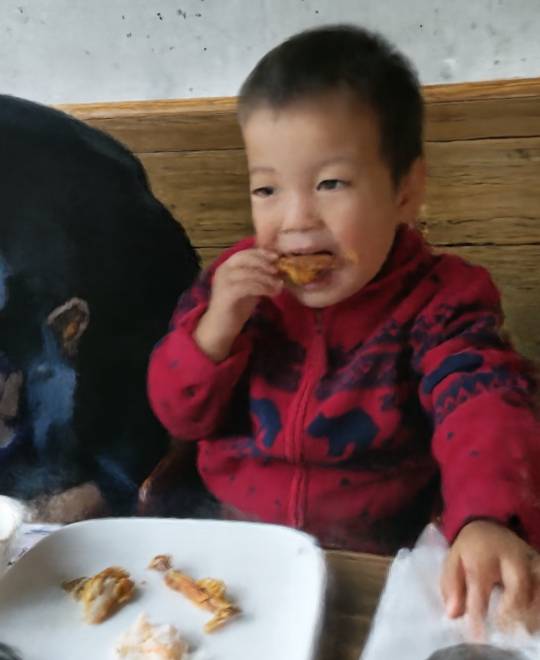}}\vspace{1mm}

\fbox{\adjincludegraphics[width=0.19\linewidth]{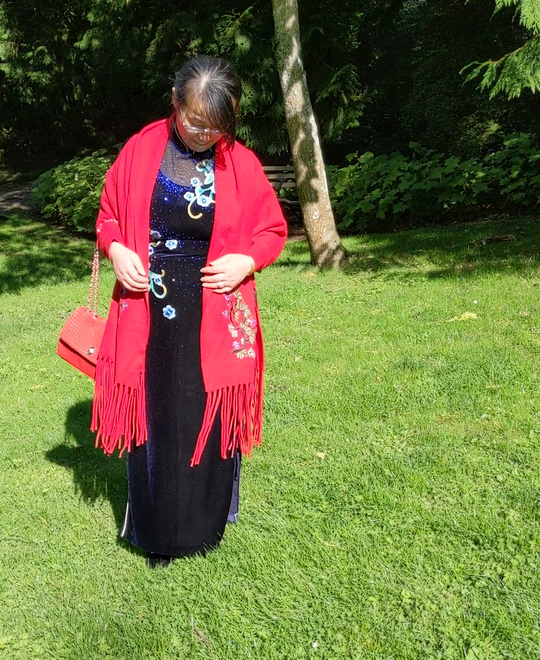}} \hfill
\fbox{\adjincludegraphics[width=0.19\linewidth]{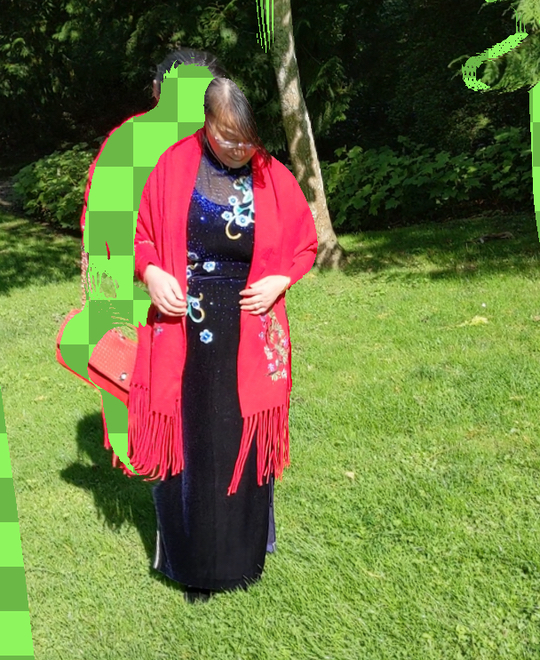}} \hfill
\fbox{\adjincludegraphics[width=0.19\linewidth]{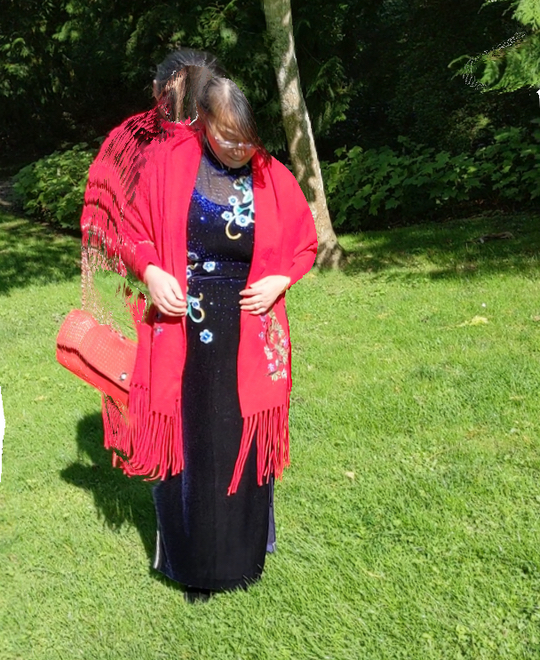}} \hfill
\fbox{\adjincludegraphics[width=0.19\linewidth]{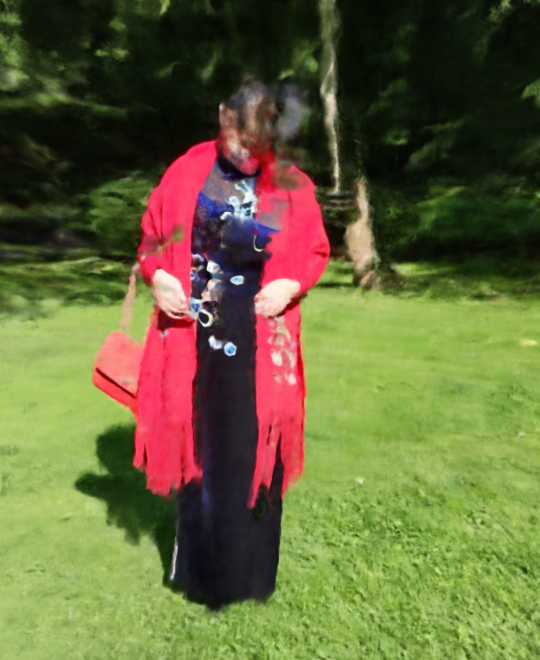}} \hfill
\fbox{\adjincludegraphics[width=0.19\linewidth]{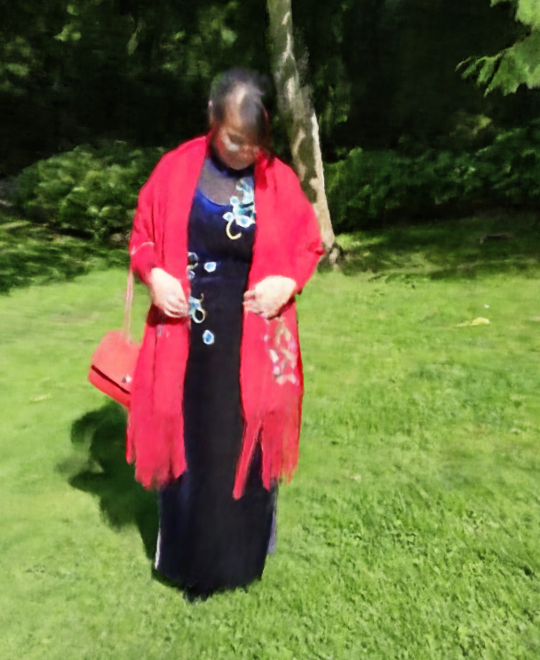}}\vspace{1mm}

\fbox{\adjincludegraphics[width=0.19\linewidth]{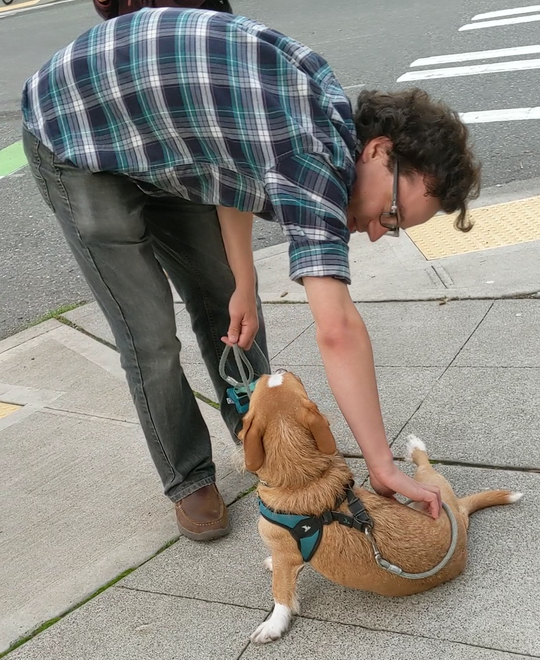}} \hfill
\fbox{\adjincludegraphics[width=0.19\linewidth]{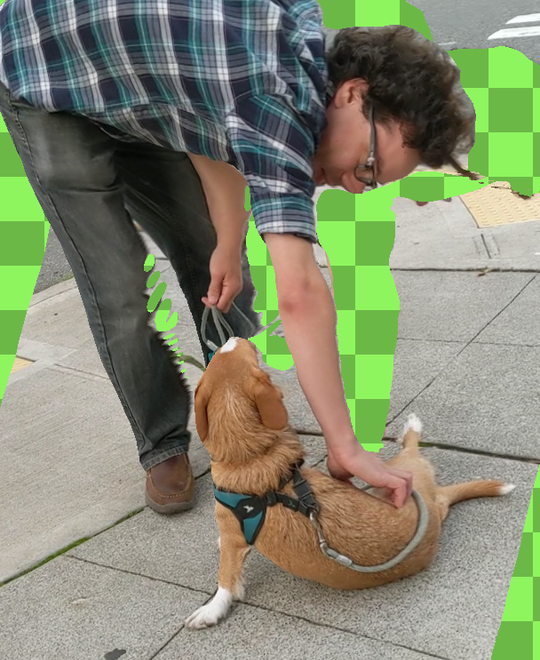}} \hfill
\fbox{\adjincludegraphics[width=0.19\linewidth]{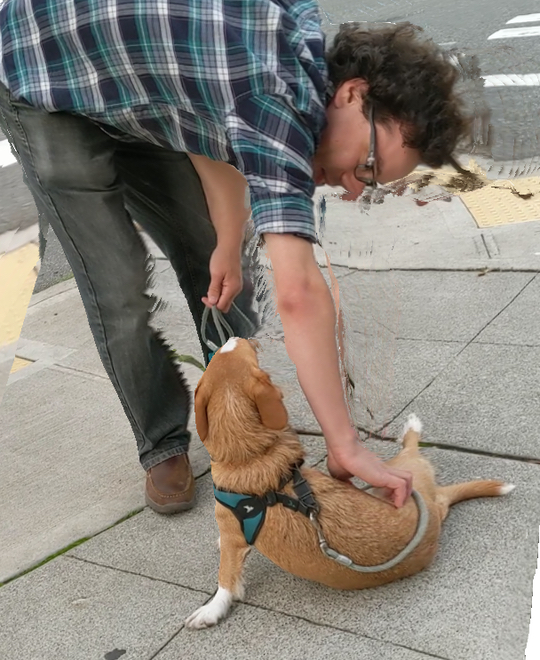}} \hfill
\fbox{\adjincludegraphics[width=0.19\linewidth]{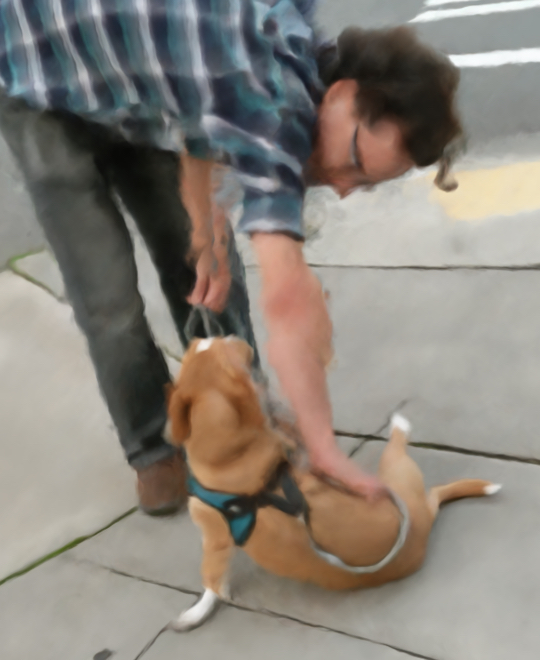}} \hfill
\fbox{\adjincludegraphics[width=0.19\linewidth]{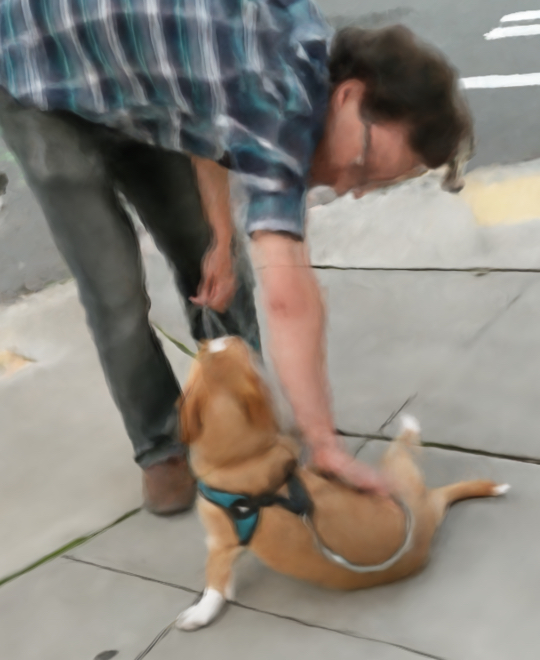}}\vspace{-1.0mm}

\mpage{0.18}{\small (a) Input} \hfill
\mpage{0.18}{\small (b) Mesh} \hfill
\mpage{0.18}{\small (c) Inpainted} \hfill
\mpage{0.18}{\small (d) NeRF-T} \hfill
\mpage{0.18}{\small (e) Ours}

\end{center}
\caption{
\tb{Comparisons on novel view synthesis.} 
We compare our results with three baselines. (a) shows an input image and (b--d) show view synthesis results. (b) shows rendered mesh representation; (c) inpainted version of (b); (d) NeRF with an additional temporal parameter; (e) our method. The green areas represent disocclusion. Notice how the baselines are unable to recover the realistic appearance that respects an accurate geometry of the background scene. Results are best appreciated in the supplementary videos. 
}
\label{fig:eval}
\end{figure*}

\topic{Implementation details.}
%
%
We used the same MLP architecture as in NeRF~\cite{mildenhall2020nerf}, except that we use 1024 activations for the first 8 layers instead of 256. We calculate all the losses except the static scene loss on a batch of $\NumRays = 1024$ rays that are randomly drawn from an input frame $\I_\T$ without replacement. We normalize the time $\T$ such that $\Tdom = [-1, 1]$ and apply the positional encoding with 4 frequency bands. We use the hierarchical volume sampling as in the original NeRF and simultaneously train both the coarse and fine networks. We train the MLP for 800k iterations. It takes about 48 hours to train a network with about 100 video frames at the 960$\times$540 resolution with 4 NVIDIA V100 GPUs. Please refer to the supplementary document for more details.

\begin{table*}
\begin{center}
\newcommand{\V}{\checkmark}
\newcommand{\N}{\color{gray}-}
\newcommand{\psnrf}{\parbox{.9cm}{\centering PSNR\\(All)}}
\newcommand{\psnro}{\parbox{.7cm}{\centering PSNR\\(Occ.)}}
\newcommand{\ssim}{\parbox{.9cm}{\centering SSIM\\(All)}}
\newcommand{\bb}[1]{\textbf{#1}}
\newcommand{\uu}[1]{\underline{#1}}
\caption{
\tb{Ablation study.}
We trained our model on the synthetic ``Sintel'' dataset with varying combinations of our losses and measure the view synthesis quality using three metrics: (1) ``PSNR (All)'' the PSNR on all pixels, (2) ``PSNR (Occ.)'' the PSNR on disoccluded pixels only, and (3) ``SSIM'' the structural similarity. With all metrics, the higher the better. Bold faces mean the best score, and the underlined mean the second best.
}
\vspace{-2mm}
\label{tab:ablation}
\resizebox{\textwidth}{!}{%
\scriptsize
\begin{tabular}{@{}l@{}c@{}c@{}c@{}c@{}c@{}c@{}c@{}c@{}c@{}c@{}c@{}c@{}c@{}c@{}c@{}c@{}c@{}c@{}c@{}c@{}c@{}c@{}c@{}c@{}}
\toprule
\multirow{2}{*}{\parbox{.7cm}{\phantom{(a)}\\Model}} &
\multirow{2}{*}{\parbox{.7cm}{\centering \phantom{(a)}\\Time}} &
\multirow{2}{*}{\parbox{.7cm}{\centering \phantom{(a)}\\$\Ldep$\\\eqref{eq:depth_loss}}} &
\multirow{2}{*}{\parbox{.7cm}{\centering \phantom{(a)}\\$\Lfree$\\\eqref{eq:zvd_loss}}} &
\multirow{2}{*}{\parbox{.7cm}{\centering \phantom{(a)}\\$\Lstatic$\\\eqref{eq:static_loss}}} &
\multirow{2}{*}{\parbox{.7cm}{\centering \phantom{(a)}\\$\Lflow$\\\phantom{(a)}}} & 
\multirow{2}{*}{\parbox{.7cm}{\centering \phantom{(a)}\\View\\dir.}} &
\multicolumn{3}{c}{``Bandage 1''} &
\multicolumn{3}{c}{``Bandage 2''} &
\multicolumn{3}{c}{``Sleeping 1''} &
\multicolumn{3}{c}{``Sleeping 2''} &
\multicolumn{3}{c}{``Alley 1''} &
\multicolumn{3}{c}{``Bamboo 1''} \\
\cmidrule(lr){8-10}
\cmidrule(lr){11-13}
\cmidrule(lr){14-16}
\cmidrule(lr){17-19}
\cmidrule(lr){20-22}
\cmidrule(lr){23-25}
&&&&&&&
\psnrf & \psnro & \ssim &
\psnrf & \psnro & \ssim &
\psnrf & \psnro & \ssim &
\psnrf & \psnro & \ssim &
\psnrf & \psnro & \ssim &
\psnrf & \psnro & \ssim \\
\midrule
NeRF   & \N & \N & \N & \N & \N & \N &    11.65 &    11.32 &    0.638 &    13.41 &    12.19 &    0.499 &    14.67 &    16.03 &    0.641 &    20.49 &    20.32 &    0.818 &    10.93 &    10.93 &    0.641 &    17.28 &    17.52 &    0.792 \\ 
NeRF-T & \V & \N & \N & \N & \N & \N &    11.81 &    11.27 &    0.658 &    13.84 &    12.26 &    0.462 &    15.45 &    15.81 &    0.663 &    13.83 &    22.08 &    0.480 &    11.50 &    12.50 &    0.665 &    15.59 &    15.32 &    0.732 \\ 
1      & \V & \V & \N & \N & \N & \N &    14.82 &    12.87 &    0.796 &    18.61 &    18.81 &    0.807 &    21.85 &    21.43 &    0.830 &    31.01 &    32.93 &    0.965 &    24.67 &    25.28 &    0.930 &\bb{28.35}&\bb{26.34}&\bb{0.966}\\ 
\midrule
2      & \V & \V & \V & \N & \N & \N &    14.62 &    13.06 &    0.791 &\bb{22.26}&\bb{23.64}&\bb{0.877}&    21.70 &    21.19 &    0.841 &\uu{35.06}&    32.61 &    0.965 &\bb{26.75}&\uu{26.40}&\bb{0.941}&    25.69 &    23.83 &\uu{0.965}\\ 
3      & \V & \V & \N & \V & \N & \N &    15.48 &    13.15 &    0.812 &\uu{20.41}&\uu{22.11}&\uu{0.860}&    21.85 &\uu{21.59}&    \uu{0.841} &\bb{35.79}&\bb{34.79}&\bb{0.982}&    25.34 &    21.42 &    0.937 &    27.20 &    23.59 &    0.957 \\ 
4 (``Ours'')      & \V & \V & \V & \V & \N & \N &    16.34 &    \uu{15.54} &   \uu{ 0.844} &    20.25 &    21.99 &    0.855 &\bb{22.53}&\bb{21.65}&\bb{0.844}&    35.01 &\uu{33.91}&\uu{0.981}&\uu{26.56}&\bb{26.75}&    0.931 &\uu{27.87}&\uu{25.28}&    0.964 \\ 
\midrule
5      & \V & \V & \N & \V & \V & \N &    \uu{17.24} &\bb{16.84}&    0.826 &    18.67 &    16.75 &    0.820 &    21.96 &    21.19 &    0.837 &    28.43 &    27.21 &    0.957 &    25.37 &    25.24 &\uu{0.940}&    25.45 &    22.97 &    0.924 \\ 
6      & \V & \V & \V & \N & \V & \N &    17.24 &    14.50 &    0.832 &    17.07 &    16.59 &    0.772 &\uu{22.37}&    21.57 &    0.841 &    28.60 &    26.20 &    0.957 &    21.65 &    22.60 &    0.920 &    21.30 &    18.62 &    0.895 \\ 
7     & \V & \V & \V & \V & \V & \V &\bb{18.14}&    14.02 &\bb{0.855}&    18.77 &    18.32 &    0.810 &    18.36 &    18.95 &    0.808 &    28.26 &    26.71 &    0.944 &    20.05 &    24.05 &    0.899 &    20.70 &    18.08 &    0.885 \\ 
\bottomrule
\end{tabular}
}
\end{center}
\vspace{-6mm}
\end{table*}

\section{Experimental Results}
\label{sec:results}

\noindent
We first compare our method with baseline approaches using the videos of dynamic scenes. 
Note that we always render new views at one of the observed times. 
That is, we do not evaluate our method's capability of temporal interpolation/extrapolation since our method is not designed to address it. 
We then provide extensive quantitative ablation studies with the variants of our model where each loss is added one at a time. 
We urge the readers to watch our supplementary videos in \emph{our project webpage} (\url{https://video-nerf.github.io}), where we provide free-viewpoint rendering of our learned representations.


\topic{Datasets.}
We use the videos of dynamic scenes from the recent consistent video estimation method of Xuan et al.~\cite{Luo-VideoDepth-2020} along with the camera calibration and the per-frame depth maps provided together.
\footnote{We also intended to use the dataset and video depth of Yoon et al.~\cite{Yoon-2020-CVPR}, but were unable to obtain the depth maps used in their results.}
Their dataset (denoted by ``CVD'') consists of short videos of moving subjects captured by a smartphone.

For quantitative evaluation, we use the synthetic stereo videos from the MPI Sintel dataset~\cite{Butler12}. 
We select 6 videos that show a variety of characteristics in terms of scene motion, camera motion, and the size of moving subjects. 
We use the left video to train our models and render them from the right video's viewpoints for ground truth comparisons.

\topic{Baselines.}
We compare our method against several baseline methods: 
(``Mesh'') textured mesh representations directly reconstructed from the input depth maps as demonstrated by Xuan et al.~\cite{Luo-VideoDepth-2020}; 
(``Inpainted'') its inpainted version, where disoccluded empty pixels in the 2D rendered images are inpainted using a recent video inpainting method~\cite{gao2020flow}; and 
(``NeRF-T'') a version of NeRF with an extra time parameter, which is our model trained with only the color reconstruction loss~\eqref{eq:color_loss}.
Note that we do \emph{not} use the viewing directions for the NeRF baseline. 

\topic{Qualitative comparisons.}
Figure~\ref{fig:eval} presents the comparisons of our model against the baselines using the ``CVD'' dataset, where we show view synthesis results from novel viewpoints.
We used our full method with all losses presented in Section~\ref{sec:overview} to create these results.
Please refer to \emph{our project webpage} for the full video results.


\topic{Ablation studies.}
We trained our model with the different combinations of losses and measured the view synthesis quality in three metrics, PSNR on the entire image, PSNR on the disoccluded regions only, and SSIM on the entire image.
Table~\ref{tab:ablation} summarizes the results.
In addition to the four losses presented in Section~\ref{sec:overview}, we test an addition loss, $\Lflow$, which measures the consistency of the color and the volume density between two corresponding spatiotemporal locations via \emph{scene flow}. 
We obtain the scene flow from the 2D optical flow raised to 3D using the per-frame scene depth. 
Our motivation is to further encourage temporal smoothness and disocclusion handling. 
Since we are effectively gather view-dependent appearances with the scene flow loss, we tested both models with and without the viewing directions used.

For quantitative evaluation, we train our models using the Sintel dataset. 
The left videos are used to train the models rendered from the right videos' viewpoints and then compared to the ground truth right videos. 
All metrics aggregate the scores over all frames.
Our model always works better when trained with the depth loss. While varying depending on scene types, the static loss and empty space loss help improve the results as well. 
However, we find that the scene flow loss does not help improve quality. 
We suspect that this is because casual videos often include strong image-space aberration such as exposure or color changes and monocular depth estimates does not provide as accurate scene depth as stereo-based methods do.
For example, this could be addressed by a latent code factoring out such variations, similarly done in NeRF-W~\cite{martinbrualla2020nerfw}.
While our ``Model-2'' works slightly better than our full model (``Model-4'') quantitatively, we have found that our full model works usually the best for real data.










\section{Conclusions}
\label{sec:conclusions}

\noindent
We have presented a simple yet effective algorithm for learning space-time irradiance fields from single casually captured videos. Our core technical contributions are (1)~leveraging monocular video depth estimation to constrain the time-varying geometry of our learned neural implicit functions and 
(2)~designing a static scene loss and a sampling strategy to propagate scene contents across time.
We extensively validate and justify our design choices both visually and quantitatively on the Sintel dataset.
We showcase free-viewpoint video rendering of several challenging dynamic scenes captured with hand-held cellphone cameras.

\paragraph{Acknowledgments.}
We thank Ayush Saraf for his help with distributed training.
All photos of individuals in this paper are used with permission.

{\small
\bibliographystyle{ieee_fullname}
\bibliography{main}
}
\newpage
\appendix

\section{Additional Details}
\noindent
In this appendix, we provide implementation details (in Section~\ref{sec:impl_detail}), training details (in Section~\ref{sec:train_detail})  and additional quantitative comparison in Section~\ref{sec:comp_mesh}. We include additional qualitative results in our project website \url{https://video-nerf.github.io}. In particular, we test on 10 different videos and provide comparison with baselines and different loss configurations.

\subsection{Implementation Details}
\label{sec:impl_detail}
\noindent
We implement our framework using PyTorch. In all our experiments, we empirically set the hyper-parameters as $\alpha = 1$, $\beta = 100$, and $\gamma = 10$. 

We calculate all the losses except the static scene loss on a batch of $\NumRays = 1024$ rays that are randomly drawn from an input frame $\I_\T$ without replacement. 
We randomly choose $\NumStaticSamples = 1024$ from $\pool$ at each step (also without replacement) for the static scene loss.
We normalize the time $\T$ such that $\Tdom = [-1, 1]$ and apply the positional encoding with 4 frequency bands.
Following NeRF~\cite{mildenhall2020nerf}, we apply positional encoding to spatial positions $\X$.
While we do not use the normalized device coordinates, we sample each ray uniformly in \emph{inverse} depth. 
We set the depth range $\Zn$ and $\Zf$ as the global minimum and maximum of all frames' depth values.

\subsection{Training Details}
\label{sec:train_detail}
\noindent
We used the same MLP architecture as in~\cite{mildenhall2020nerf}, except
that we use 1024 activations for the first 8 MLP layers instead of 256. Our models are trained with various combinations of the losses presented in the main paper but otherwise with the same hyperparameters.
We used the Adam optimizer with momentum parameters $\beta_1 = 0.9$ and $\beta_2 = 0.999$ and a learning rate of 0.0005. 
We train the MLP for 800k iterations. 
It takes about 48 hours to train a network with about 100 video frames at the 960$\times$540 resolution with 4 NVIDIA V100 GPUs.

\subsection{Additional Comparisons to Baseline}
\label{sec:comp_mesh}
\begin{table}[b]
\begin{center}
\caption{Reported PSNR on disoccluded pixels only.}
\resizebox{\columnwidth}{!}{%
\begin{tabular}{@{}lcccccc@{}} 
\toprule
Methods
& Bandage1 & Bandage2 & Sleeping1 & Sleeping2 & Alley1 & Bamboo1 \\ \midrule

Inpainted  &   \textbf{16.52}  &  13.32   &   20.08   &   17.52   &    14.82     &  12.23                     \\
Ours  &   15.54  &    \textbf{21.99}   &  \textbf{21.65}   &   \textbf{33.91}  &   \textbf{26.75}   &   \textbf{25.28}  \\
\bottomrule
\end{tabular}%
}
\label{tab:table1}
\end{center}
\end{table}

\noindent
In this section, we provide a quantitative comparison to the inpainted mesh method. We rendered the ground-truth mesh provided by the Sintel dataset in 2D and inpaint the missing pixels in disoccluded areas using state-of-the-arts video completion algorithms. Then, we evaluate how well our method handles disoccluded areas, compared with the inpainted mesh method, for which we used the same Sintel GT depth. Table~\ref{tab:table1} shows the quantitative evaluation measured in PSNR metric.
It demonstrates that inpainting is not sufficient to get good disocclusion contents, and validates that our approach produces significantly fewer artifacts than the baseline method using video completion.

\end{document}